\documentclass[10pt,twocolumn,letterpaper]{article}

\usepackage[pagenumbers]{wacv} %
\usepackage[accsupp]{axessibility}
\usepackage{graphicx}
\usepackage{amsmath}
\usepackage{amssymb}
\usepackage{booktabs}
\usepackage{colortbl}
\usepackage{multicol}
\usepackage{multirow}
\usepackage{xcolor}
\usepackage{float}
\usepackage{amssymb}%
\usepackage{pifont}%
\newcommand{\cmark}{\ding{51}}%
\newcommand{\xmark}{\ding{55}}%
\usepackage[pagebackref,breaklinks,colorlinks]{hyperref}

\usepackage[capitalize]{cleveref}
\crefname{section}{Sec.}{Secs.}
\Crefname{section}{Section}{Sections}
\Crefname{table}{Table}{Tables}
\crefname{table}{Tab.}{Tabs.}

\def\wacvPaperID{378} %
\def\confName{WACV}
\def\confYear{2025}

\begin{document}
\newcommand*\samethanks[1][\value{footnote}]{\footnotemark[#1]}

\title{Domain Generalization using Large Pretrained Models with Mixture-of-Adapters}

\author{Gyuseong Lee$^1$\thanks{Equal Contribution} \qquad Wooseok Jang$^1$\samethanks \qquad Jinhyeon Kim$^1$ \qquad Jaewoo Jung$^2$ \qquad Seungryong Kim$^2$\thanks{Corresponding author.}\\
$^1$Korea University \qquad $^2$KAIST\\
}
\maketitle

\begin{abstract}
Learning robust vision models that perform well in out-of-distribution (OOD) situations is an important task for model deployment in real-world settings. Despite extensive research in this field, many proposed methods have only shown minor performance improvements compared to the simplest empirical risk minimization (ERM) approach, which was evaluated on a benchmark with a limited hyperparameter search space. Our focus in this study is on leveraging the knowledge of large pretrained models to improve handling of OOD scenarios and tackle domain generalization problems. However, prior research has revealed that naively fine-tuning a large pretrained model can impair OOD robustness. Thus, we employ parameter-efficient fine-tuning (PEFT) techniques to effectively preserve OOD robustness while working with large models. Our extensive experiments and analysis confirm that the most effective approaches involve ensembling diverse models and increasing the scale of pretraining. As a result, we achieve state-of-the-art performance in domain generalization tasks. 
Our code and project page are available at: \href{https://cvlab-kaist.github.io/MoA}{\texttt{https://cvlab-kaist.github.io/MoA}}
\end{abstract}

\section{Introduction}
\label{sec:intro}
\begin{figure}[t]
\begin{center}
    \includegraphics[width=0.4\textwidth]{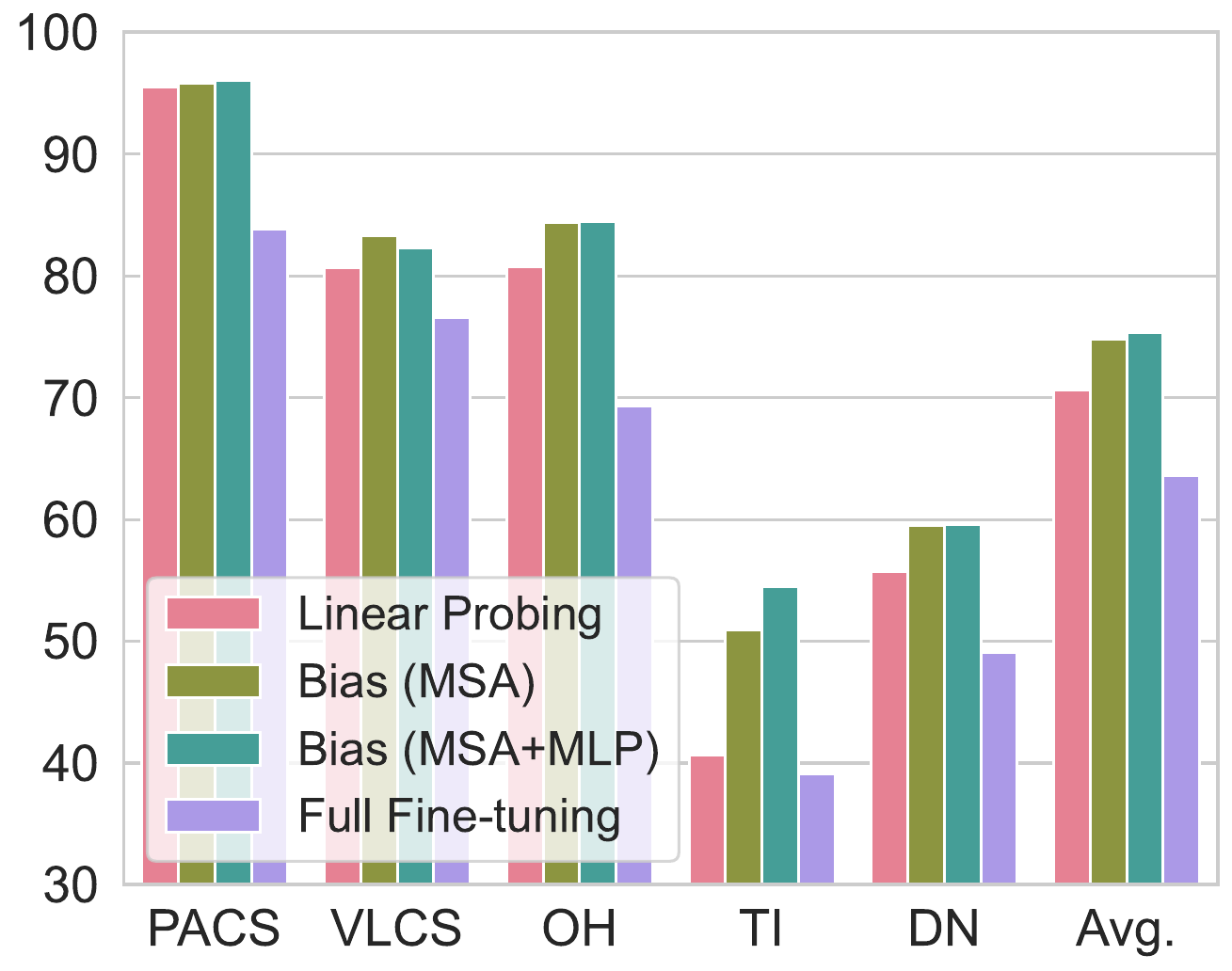}
\end{center} \vspace{-10pt}
\caption{Results on domain generalization benchmarks with varying trainable parameters in ViT-B/16~\cite{dosovitskiy2020image}, pretrained on a private OpenAI dataset~\cite{radford2021learning}. The y-axis indicates accuracy. We use linear probing (denoted as Linear), bias tuning in the attention layer (Bias (MSA)), bias tuning in both the attention and MLP layers (Bias (MSA+MLP)), and full fine-tuning to illustrate how accuracy changes with different trainable parameters when applying PEFT methods to large models. OH, TI, DN denotes OfficeHome~\cite{venkateswara2017deep}, TerraIncognita~\cite{beery2018recognition}, and DomainNet~\cite{peng2019moment}, respectively.} \vspace{-10pt}
\label{fig:motivation}
\end{figure}

\begin{figure*}[t]
    \centering
    \begin{subfigure}[t]{0.24\textwidth}
        \includegraphics[width=\textwidth]{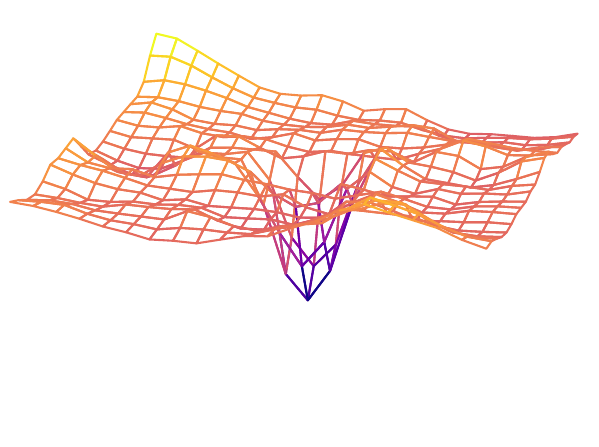}
        \caption{Fully fine-tuned}
        \label{fig:loss_surface_fullft}
    \end{subfigure}
    \begin{subfigure}[t]{0.24\textwidth}
        \includegraphics[width=\textwidth]{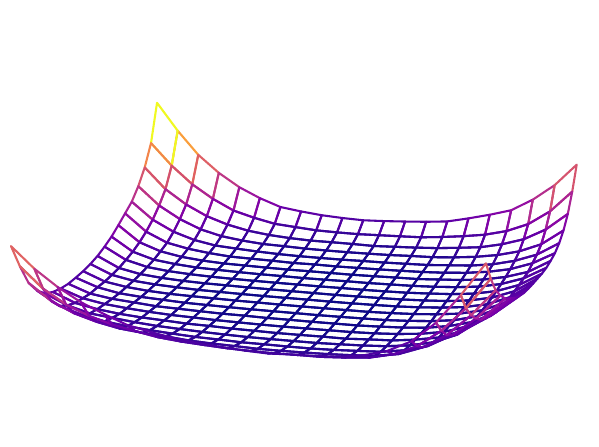}
        \caption{LoRA}
        \label{fig:loss_surface_lora}
    \end{subfigure}
    \begin{subfigure}[t]{0.24\textwidth}
        \includegraphics[width=\textwidth]{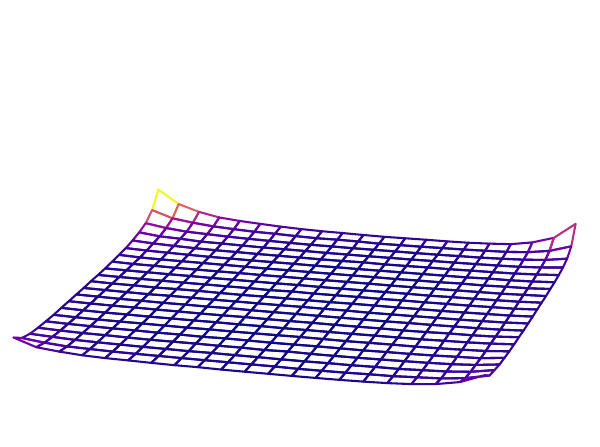}
        \caption{KAdaptation}
        \label{fig:loss_surface_ka}
    \end{subfigure}
    \begin{subfigure}[t]{0.24\textwidth}
        \includegraphics[width=\textwidth]{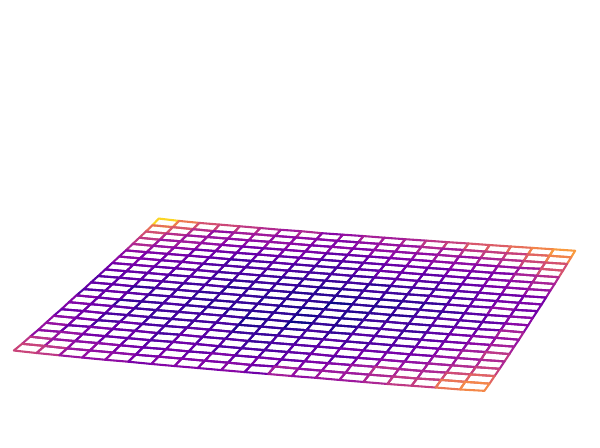}
        \caption{KMoA (Ours)}
    \end{subfigure}
    \caption{Flatness comparison of loss surfaces from models trained with full fine-tuning, LoRA, KAdaptation, and KAdaptation with Mixture-of-Adapter (denoted as KMoA) on PACS dataset~\cite{li2017deeper}. All visualizations are computed from test environment 0 (Art) domain. The x and y axes (plane) in each figure represent the perturbation directions of the model weight, and the z axis (height) represents the change in loss value according to the weight perturbation.}\vspace{-10pt}
    \label{fig:loss_surface}
\end{figure*}

Domain generalization (DG) aims to learn generalizable features from different domains to perform well on other domains that were unavailable during training (a.k.a. unseen domains)~\cite{gulrajani2020search}. Although each domain shares the same class information, their visual appearances differ. For example, a bird can look quite different in sketches, clip-art, and real photos. Therefore, the model must be able to identify domain-invariant features of the bird to generalize effectively to completely new, unseen domains. Standard benchmarks for DG~\cite{gulrajani2020search} involve training the model on multiple source domains and evaluating it on one or more unseen target domains. Due to this constraint, DG algorithms~\cite{vapnik1998statistical,arjovsky2019invariant,sagawa2019distributionally,sun2016deep,li2018domain,ganin2016domain} are developed to extract domain-invariant features from multiple source domains. For example, \cite{seo2020learning} designs network architectures to learn domain-agnostic representations, while \cite{arjovsky2019invariant} constructs a feature representation in a way that guarantees the uniform alignment of the optimal linear classifier across different domains.

Recently, large pretrained models have become increasingly popular in the field of domain generalization. \cite{cha2022domain} introduces the concept of an oracle model, which is designed to generalize across all domains and functions as an approximation of a large pretrained model. \cite{mao2022context, li2023simple,cha2022domain} leverages large pretrained models such as CLIP~\cite{radford2021learning} and SWAG~\cite{singh2022revisiting}, which offer richer and more effective domain-generalized features. As these models already contain a degree of domain-invariant knowledge~\cite{cha2022domain}, harnessing this knowledge for domain generalization has gained popularity. Despite its simplicity and efficiency, we find that only a few studies have attempted to train these models directly using empirical risk minimization (ERM)~\cite{vapnik1998statistical}. \cite{angarano2022back} shows that the ERM algorithm performs competitively well when accompanied with proper backbones like EfficientNet~\cite{tan2019efficientnet}, ViT~\cite{dosovitskiy2020image}, DeiT~\cite{touvron2021training}, LeViT~\cite{graham2021levit} and ConViT~\cite{d2021convit}. This inspired us to use large pretrained models within DG settings.

We propose to adopt parameter-efficient fine-tuning (PEFT) methods in the context of DG to leverage large pretrained models, thereby extending their application beyond traditional transfer learning domains~\cite{houlsby2019parameter}. PEFT methods mitigate the high cost associated with fully fine-tuning large pretrained models~\cite{he2022parameter, chen2022vision}. They aim to achieve or exceed the performance of full fine-tuning or zero-shot learning by tuning only specific parts of the model~\cite{zaken2021bitfit} or by adding a small number of learnable parameters, such as those used in LoRA~\cite{hu2021lora} or Compacter~\cite{karimi2021compacter}.
We tested various trainable parameter settings, including full fine-tuning and adapter fine-tuning. Our empirical results show that PEFT methods effectively act as regularization during training. These methods not only significantly outperform fully fine-tuned models but also achieve performance comparable to the recent state-of-the-art methods SWAD~\cite{cha2021swad}, MIRO~\cite{cha2022domain}, EoA~\cite{arpit2022ensemble}, and SIMPLE~\cite{li2023simple} in DG settings.

Many studies have pointed out that simply fine-tuning all parameters causes the model to \textit{forget} the knowledge acquired during the pretraining phase and to lose its robustness~\cite{kumar2022fine,wortsman2022robust}. Therefore, by regularizing the trainable parameters, we can mitigate the forgetting problem~\cite{touvron2022three,zaken2021bitfit,hu2021lora} and achieve better results compared to the naive full fine-tuning strategy. As shown in Fig.~\ref{fig:motivation}, the fully fine-tuned CLIP model shows a significant performance drop on many datasets. In contrast, the most restrictive setting, which involves freezing nearly all the trainable parameters and training only the last linear layer (Linear Probing), demonstrates much better performance on all datasets except for TerraIncognita~\cite{beery2018recognition}. We observe that datasets with significant visual differences from traditional ones, such as the TerraIncognita (TI) dataset, which contains images of wild animals captured by automatic cameras, require different levels of regularization compared to traditional datasets. For example, on TI, Bias (MSA+MLP) significantly outperforms both Bias (MSA) and the Linear method. We argue that the lower performance of the Bias (MSA) and Linear methods on TI is caused by excessive regularization, resulting from their relatively smaller number of tunable parameters compared to the Bias (MSA+MLP) method.

These findings highlight the importance of carefully adjusting the amount of regularization through trainable parameters to effectively manage varying distribution shifts. This prompts the question: \textit{Can we determine the optimal model capacity by tuning the trainable parameters?} To address this challenge, we introduce a mixture-of-experts-based adapter architecture called Mixture-of-Adapters (MoA). This approach effectively adjusts regularization strength using adapters with varying capacities, directing tokens to the most suitable ones. By leveraging MoA and learnable routers, it handles distribution shifts across datasets, achieving state-of-the-art results on DG benchmarks.

\section{Related Work}

\subsection{Domain Generalization}
For the past decade, numerous learning methods on how to learn domain invariant representations have been proposed in the domain generalization field. Empirical risk minimization (ERM)~\cite{vapnik1998statistical} which is one of the most simplest approaches, just minimizes the loss on each domain and trains the model. In DomainBed~\cite{gulrajani2020search}, ERM still remains effective within the benchmark's restricted hyperparameter search space and model selection methods. DomainBed tested many DG methods such as IRM~\cite{arjovsky2019invariant}, GroupDRO~\cite{sagawa2019distributionally}, MixUp~\cite{xu2020adversarial}, DANN~\cite{ganin2016domain}, and CORAL~\cite{sun2016deep} in a unified and contained experimental setup. 
SWAD~\cite{cha2021swad} explored the relationship between flat loss surfaces and DG performance, achieving superior results. Ensemble-of-Averages (EoA)~\cite{arpit2022ensemble} perform model averaging in training time and ensemble them in test time. Current research is focused on leveraging knowledge from large pretrained models, with MIRO~\cite{cha2022domain} introducing an oracle model approximated by a large pretrained model to maximize the mutual information between the target model.
Recently, methods that ensemble diverse models show remarkable performance in the DG benchmark. SIMPLE~\cite{li2023simple} utilizes many different pretrained models from a ModelPool and extracts outputs from the frozen pretrained models, and trains a shallow dispatcher using these outputs.

\subsection{Parameter efficient fine-tuning}
Fine-tuning large pretrained models for specific tasks often involves adjusting all parameters, which can be impractical. Parameter-efficient fine-tuning (PEFT) addresses this by freezing most parameters and optimizing only a few for the task. Many successful PEFT approaches such as Bitfit~\cite{zaken2021bitfit}, LoRA~\cite{hu2021lora}, LM-BFF~\cite{gao2020making}, CaFo~\cite{zhang2023prompt}, Pointclip~\cite{zhang2022pointclip}, VL-Adapter~\cite{sung2022vl}, LST~\cite{sung2022lst} adopt popular pretrained models to various downstream tasks. Among these methods, the use of adapters are adopted because of its high performance and efficient computation cost. Adapters are small modules trained on specific tasks which are inserted between network layers, where all the layers except for the adapters are frozen during training. Adapters such as LoRA, Compacter~\cite{karimi2021compacter} and KAdaptation~\cite{he2022parameter} greatly reduced the number of trainable parameters by attaching a low-rank hypercomplex adapter layers in the transformer model and by decomposing the updated weight matrix into low rank matrices respectively. 
\begin{figure}[t]
    \centering
    \begin{subfigure}[t]{1.0\linewidth}
        \begin{subfigure}[t]{0.49\linewidth}
            \includegraphics[width=\linewidth]{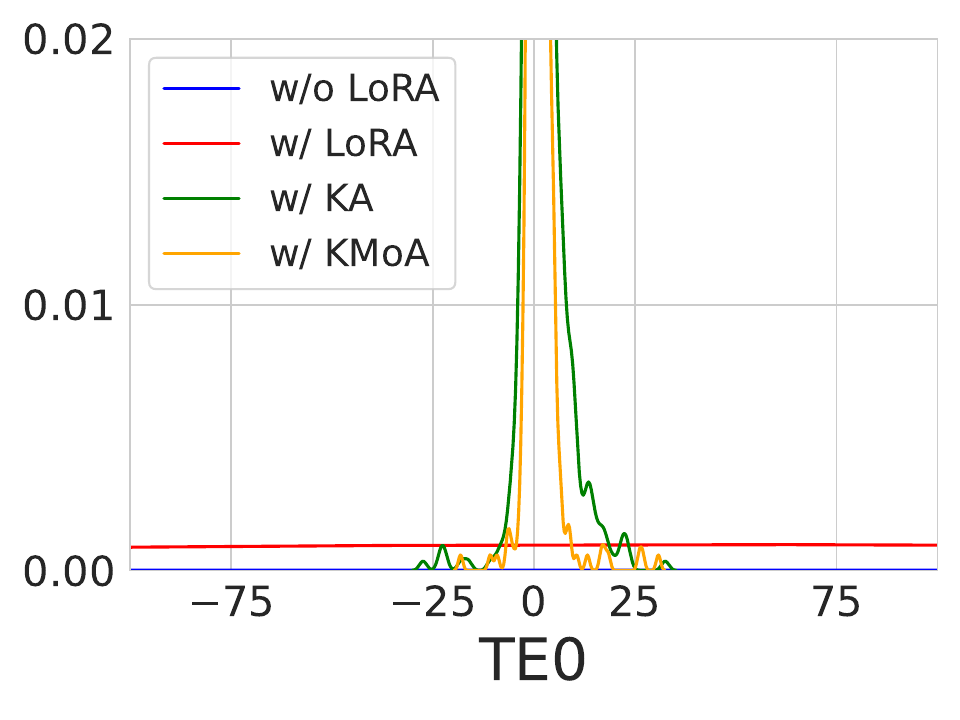}
        \end{subfigure}
        \begin{subfigure}[t]{0.49\linewidth}
            \includegraphics[width=\linewidth]{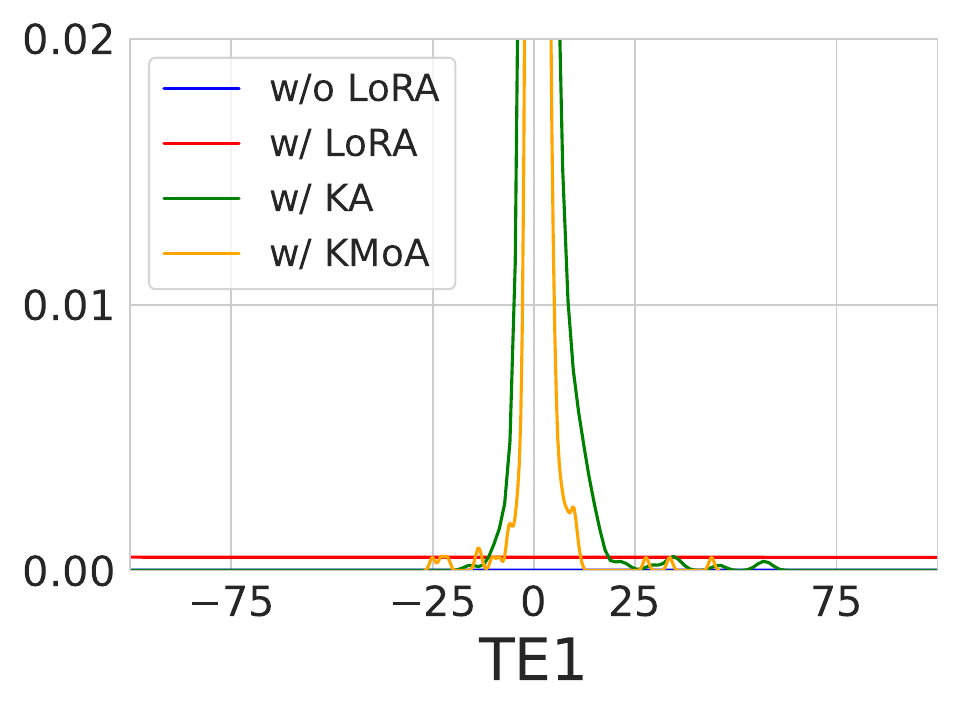}
        \end{subfigure}
    \end{subfigure}\\
    \begin{subfigure}[t]{1.0\linewidth}
        \begin{subfigure}[t]{0.49\linewidth}
            \includegraphics[width=\linewidth]{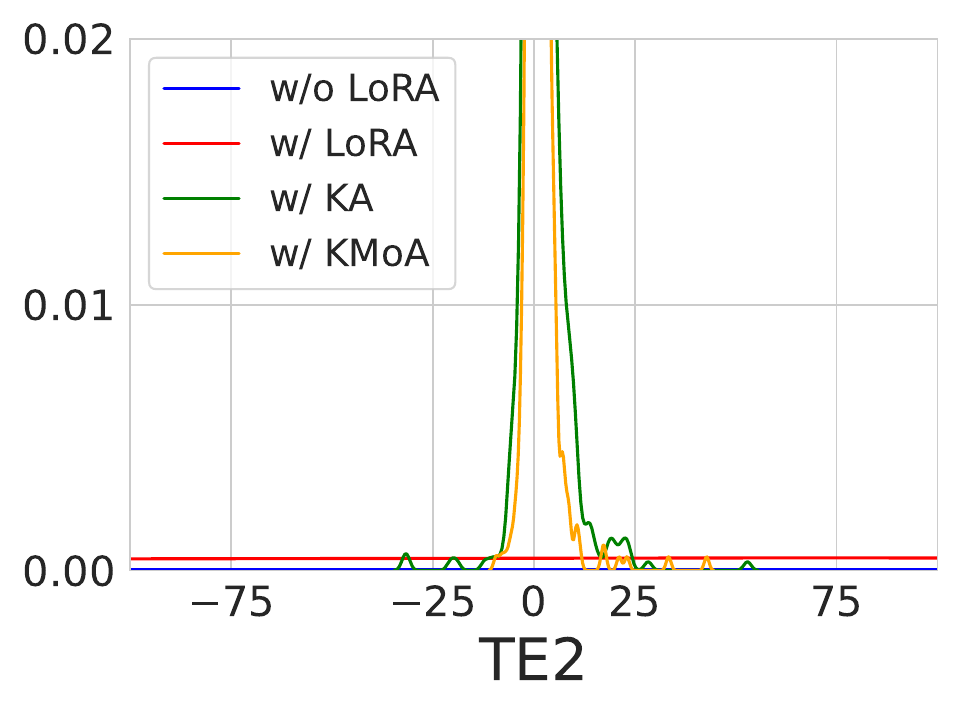}
        \end{subfigure}
        \begin{subfigure}[t]{0.49\linewidth}
            \includegraphics[width=\linewidth]{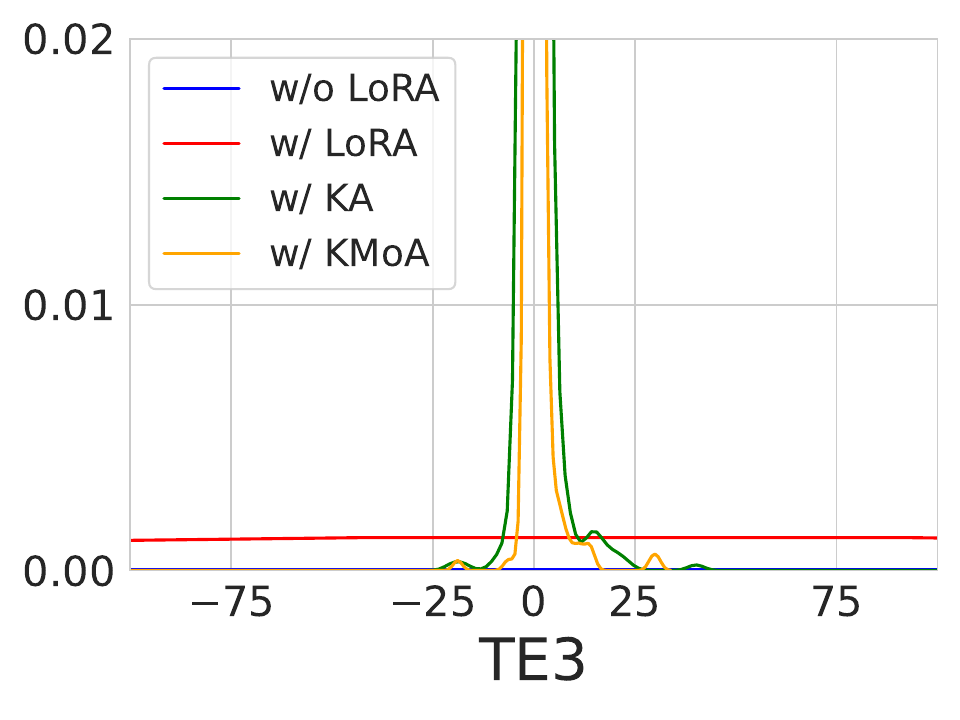}
        \end{subfigure}
    \end{subfigure}\vspace{-5pt}
\caption{A comparison of the max Hessian eigenvalue spectra trained with full fine-tuning, LoRA, KAdaptation, and MoA with KAdaptation on the PACS dataset~\cite{li2017deeper}. The x-axis represents the maximum Hessian eigenvalue, while the y-axis represents its density. KA and KMoA refer to KAdaptation and Mixture-of-Adapter with KAdaptation methods, respectively. TE0 to TE3 represent each testing domain from the domain generalization dataset. For example, in the PACS dataset, TE0 to TE3 correspond to \texttt{art\_painting}, \texttt{cartoon}, \texttt{photo}, and \texttt{sketch}, respectively.
In all test environments, KMoA exhibits the most zero-concentrated eigenvalue spectrum. KA and LoRA also show a smaller max Hessian eigenvalue distribution compared to full fine-tuning (w/o LoRA). Note that the x-axis is highly magnified, making w/ LoRA and w/o LoRA appear almost flat. However, when the x-scale is expanded, both w/ LoRA and w/o LoRA exhibit a zero-concentrated shape similar to w/ KA and w/ KMoA.} \vspace{-10pt}
\label{fig:hessian}
\end{figure}

\subsection{Mixture-of-Experts} Mixture-of-Experts (MoEs) models are proposed to improve model performance by incorporating multiple subsets of parameters called `experts' with routing algorithms conditioned by its input~\cite{jordan1994hierarchical, eigen2013learning}. Evolving from this, a type of model called sparse MoEs has become popular in both NLP (e.g., GShard~\cite{lepikhin2020gshard}, ST-MoE~\cite{zoph2022st}, GLaM~\cite{du2022glam}) and vision (e.g., CondConv~\cite{yang2019condconv}, DeepMoE~\cite{wang2020deep}, V-MoE~\cite{riquelme2021scaling}) tasks lately. This is due to its capability to enhance model capacity while simultaneously reducing the substantial increase in the computational resources required for training. In the field of DG, \cite{li2022sparse} incorporated MoE design into DG tasks, and proposed a Generalizable Mixture-of-Experts (GMoE) architecture to effectively handle distribution shifts.

\section{Preliminaries}

\subsection{Domain Generalization}
Let us denote the set of training domains as $\mathcal{D} = \{\mathcal{D}^i\}_{i=1}^{K}$ where $K$ implies the total number of training domains, and $\mathcal{D}_i$ is the distribution over the input space. Also, define the set of target domains as $\mathcal{T} = \{\mathcal{T}^i\}_{i=1}^{K'}$ where $K'$ is the number of target domains. The training dataset is composed of $n_{\mathcal{D}^i}$ data points denoted as $(x^{i}_{j},y^{i}_{j})_{j=1}^{n_{\mathcal{D}^i}}\sim\mathcal{D}^i$ where $x$ is the input and $y$ is the target label for each training domain. In DG settings, the goal is to find the model parameter $\theta$ of a classifier $f_\theta$ which generalizes well on both $\mathcal{D}$ and $\mathcal{T}$. More precisely, for the ERM algorithm, we define  
\begin{equation}
\mathcal{E}_{\mathcal{D}}(\theta)={\frac{1}{K}}\sum^{K}_{i=1}\mathbb{E}_{(x^i, y^i)\sim\mathcal{D}^i}[l(f_\theta(x^i),y^i)]
\end{equation}
where $l(\cdot,\cdot)$ denote a loss function and define $\mathcal{E}_{\mathcal{T}}$ in the similar manner. We minimize empirical risk $
\hat{\mathcal{E}}_{\mathcal{D}}(\theta)={\frac{1}{K}}\sum^{K}_{i=1}{\frac{1}{n_{\mathcal{D}^i}}}\sum^{n_{\mathcal{D}^i}}_{j=1}[l(f_\theta(x^i_j),y^i_j)]
$ during training and expect the optimal parameter $\hat{\theta}=\arg \min_{\theta}\hat{\mathcal{E}}_{\mathcal{D}}(\theta)$ to be optimal for $\mathcal{E}_{\mathcal{T}}(\theta)$.
\begin{table*}[t]
\centering
\resizebox{\textwidth}{!}{
\begin{tabular}{l|cc|ccccc|c|cc}
\toprule 
\multirow{2}{*}{Algorithm} & \multirow{2}{*}{Architecture}  & \multirow{2}{*}{Pretraining}  & \multirow{2}{*}{PACS}  & \multirow{2}{*}{VLCS}  & \multirow{2}{*}{OfficeHome}  & \multirow{2}{*}{TerraIn.}  & \multirow{2}{*}{DomainNet}  & \multirow{2}{*}{Avg.} & \multirow{2}{*}{\#Param.}  & Trainable  \\
 & & & & & & & & & & \#Param. \\ 
\midrule 
MIRO& ViT-B/16 & CLIP$_\textrm{OpenAI}$ & 95.6 \scriptsize{$\pm$ 0.8} & 82.2 \scriptsize{$\pm$ 0.3} & 82.5 \scriptsize{$\pm$ 0.1} & 54.3 \scriptsize{$\pm$ 0.4} & 54.0 \scriptsize{$\pm$ 0.3} & 73.7 & 172M & 85.8M \\
MIRO& RegNetY-16GF & SWAG$_\textrm{IG3B}$ & \underline{97.4} \scriptsize{$\pm$ 0.2} & 79.9 \scriptsize{$\pm$ 0.6} & 80.4 \scriptsize{$\pm$ 0.2} & 58.9 \scriptsize{$\pm$ 1.3} & 53.8 \scriptsize{$\pm$ 0.1} & 74.1 & 167.2M & 83.6M \\
SMA         & RegNetY-16GF & SWAG$_\textrm{IG3B}$ & 95.5 \scriptsize{$\pm$ 0.0} & 80.7 \scriptsize{$\pm$ 0.1} & 82.0 \scriptsize{$\pm$ 0.0} & \underline{59.7} \scriptsize{$\pm$ 0.0} & 60.0 \scriptsize{$\pm$ 0.0} & 75.6 & 83.6M & 83.6M \\ 
MIRO+SWAD& RegNetY-16GF & SWAG$_\textrm{IG3B}$ & 96.8 \scriptsize{$\pm$ 0.2} & 81.7 \scriptsize{$\pm$ 0.1} & 83.3 \scriptsize{$\pm$ 0.1 }& \textbf{64.3} \scriptsize{$\pm$ 0.3} & 60.7 \scriptsize{$\pm$ 0.0} & \textbf{77.3} & 167.2M & 83.6M \\  
\midrule
\multicolumn{11}{l}{\textit{Partial fine-tuning methods}} \\ \midrule
\textcolor{gray}{Linear Probing}& \textcolor{gray}{ViT-B/16} & \textcolor{gray}{CLIP$_\textrm{OpenAI}$} & \textcolor{gray}{95.5}  & \textcolor{gray}{80.7} & \textcolor{gray}{80.7}  & \textcolor{gray}{40.6}  & \textcolor{gray}{55.7} & \textcolor{gray}{70.7} & \textcolor{gray}{86.1M} & \textcolor{gray}{0.01M} \\
\textcolor{gray}{Bias (MSA)}& \textcolor{gray}{ViT-B/16} & \textcolor{gray}{CLIP$_\textrm{OpenAI}$} & \textcolor{gray}{95.8}  & \textcolor{gray}{83.3} & \textcolor{gray}{84.4}  & \textcolor{gray}{50.9}  & \textcolor{gray}{59.5} & \textcolor{gray}{74.8} & \textcolor{gray}{86.1M} & \textcolor{gray}{0.04M} \\
\textcolor{gray}{Bias (MSA + MLP)}& \textcolor{gray}{ViT-B/16} & \textcolor{gray}{CLIP$_\textrm{OpenAI}$} & \textcolor{gray}{96.1}  & \textcolor{gray}{82.3} & \textcolor{gray}{84.5}  & \textcolor{gray}{54.5}  & \textcolor{gray}{59.6} & \textcolor{gray}{75.4} & \textcolor{gray}{86.1M} & \textcolor{gray}{0.1M} \\ \midrule
\multicolumn{11}{l}{\textit{Methods with Parameter-Efficient Fine-Tuning (PEFT)}} \\
\midrule
\rowcolor{blue!10}
ERM (Baseline) & ViT-B/16 & CLIP$_\textrm{LAION2B}$ & 85.8 \scriptsize{$\pm$ 2.1} & 78.5 \scriptsize{$\pm$ 0.9}& 78.1 \scriptsize{$\pm$ 0.8} & 41.0 \scriptsize{$\pm$ 1.6} & 52.2 \scriptsize{$\pm$0.1} & 67.1 & 85.8M & 85.8M \\
\rowcolor{blue!10}
ERM$_{\textrm{LoRA},r=2}$ & ViT-B/16 & CLIP$_\textrm{LAION2B}$ & 96.4 \scriptsize{$\pm$ 0.6} & \underline{82.6} \scriptsize{$\pm$ 0.6} & \underline{86.7} \scriptsize{$\pm$ 0.3} & 46.1 \scriptsize{$\pm$1.7} & \underline{61.5} \scriptsize{$\pm$0.1} & 74.7 & 85.9M & 0.11M \\
\rowcolor{blue!10}
ERM$_\textrm{KAdaptaion}$ & ViT-B/16 & CLIP$_\textrm{LAION2B}$ & \textbf{97.5} \scriptsize{$\pm$ 0.1} & \textbf{83.0} \scriptsize{$\pm$ 0.1} & \textbf{90.3} \scriptsize{$\pm$ 0.1} & 51.9 \scriptsize{$\pm$0.5} & \textbf{62.7} \scriptsize{$\pm$0.0} & \underline{77.1} & 85.9M & 0.14M \\
\bottomrule

\end{tabular}
} \vspace{-5pt}
\caption{Quantitative evaluation on domain generalization datasets using different PEFT methods. Following DomainBed~\cite{gulrajani2020search}, we evaluate our algorithm on PACS~\cite{li2017deeper}, VLCS~\cite{fang2013unbiased}, OfficeHome~\cite{venkateswara2017deep}, TerraIncognita~\cite{beery2018recognition} and DomainNet~\cite{peng2019moment}. To ensure a fair comparison, we use the training-domain-validation model selection strategy. The performances of each algorithm are sourced from their respective papers. %
The pretraining method and dataset are indicated as $\mathtt{PRETRAIN}_\mathtt{DATASET}$. OpenAI refers to the private dataset used to train CLIP~\cite{radford2021learning}, 
and IG3B refers to the Instagram-3B dataset from \cite{zellers2018swag}.
Results obtained with our method are highlighted in blue.}\label{tab:peft-compare}
\vspace{-10pt}
\end{table*}

\subsection{Parameter Efficient Adapter}
Parameter-efficient fine-tuning (PEFT) efficiently adapts large pretrained models to downstream tasks and achieves significant performance, unlike direct fine-tuning methods that require high memory and computational costs. Previous approaches to fine-tuning large models have used partial fine-tuning~\cite{touvron2022three}. However, recent studies \cite{houlsby2019parameter, hu2021lora, karimi2021compacter} demonstrate that adding a learnable layer to the frozen pretrained weights can outperform partial fine-tuning methods. More specifically, \cite{aghajanyan2020intrinsic} verifies that dense neural network layers with full-rank matrices can be reduced to lower-rank subspaces. From this finding, \cite{hu2021lora} constrains the updated weight $\Delta \mathbf{W}=\mathbf{BA}\in \mathbb{R}^{k \times d}$ to have a low intrinsic rank which can be expressed as $\mathbf{h} = \mathbf{W_0}\mathbf{x} + \Delta \mathbf{W}\mathbf{x} = \mathbf{W_0}\mathbf{x} + \mathbf{BA}\mathbf{x} $, where $\mathbf{x}$ denotes input sequences of each module, $\mathbf{W_0} \in \mathbb{R}^{k \times d}$ denotes the frozen weight matrix of a pretrained model, and $\mathbf{A}\in \mathbb{R}^{k \times r}$, and $\mathbf{B}\in \mathbb{R}^{r \times d}$ $(r<k,d)$ are the trainable parameters during training. 

In addition, KAdaptation~\cite{he2022parameter} and Compacter~\cite{karimi2021compacter} both exploit Kronecker products to decompose the weight matrices into reduce trainable parameters. The Kroneker product between matrix $\mathbf{A} \in \mathbb{R}^{m \times n}$ and $\mathbf{B} \in \mathbb{R}^{p \times q}$ is denoted by $\mathbf{A} \otimes \mathbf{B} \in \mathbb{R}^{mp \times nq }$ and can be expressed as the following:
\begin{equation}\mathbf{A} \otimes \mathbf{B} = 
    \begin{pmatrix}
        a_{11}\mathbf{B}    & \textbf{\dots}  &  a_{1n}\mathbf{B} \\
        \vdots              & \ddots &  \vdots        \\    
        a_{m1}\mathbf{B}    & \textbf{\dots}  &  a_{mm}\mathbf{B} \\
    \end{pmatrix}     
\end{equation}
where $a_{ij}$ indicates the element in the $i$-th row and the $j$-th column of $\mathbf{A}$.
It decomposes the update weight matrix $\Delta\mathbf{W} = \sum_{i=1}^{t} \mathbf{A}_{i} \otimes \mathbf{B}_{i} \in \mathbb{R}^{k \times d}$ where $t$ is a hyperparameter that decides the number of Kronecker products. Furthermore, $ \mathbf{W}$ can be expressed as the following: 
$ \mathbf{W} = \sum_{i=1}^{t} \mathbf{A}_{i} \otimes \mathbf{B}_{i} = \sum_{i=1}^{n} \mathbf{A}_{i} \otimes (\mathbf{u}_i \mathbf{v}_i^\top)$
where slow-weights $\mathbf{A}_{i}  \in \mathbb{R}^{t \times t}$ is shared across all layers, and fast-weights $\mathbf{B}_{i} \in \mathbb{R}^{ \frac{k}{t} \times \frac{d}{t} }$ is decomposed into low-rank matrices $\mathbf{u}_i \in \mathbb{R}^{ \frac{k}{t} \times r}$ and $\mathbf{v}_{i} \in \mathbb{R}^{r \times \frac{d}{t} }$ with $ i \in \{1 , \dots , t\}$. Compacter decomposes the weight matrix to the additional adapter layer whereas KAdaptation decomposes the update matrix $\Delta \mathbf{W}$ in the original layer.

\begin{table*}[!t]
\centering
\resizebox{\textwidth}{!}{
\begin{tabular}{l|cc|ccccc|c|cc}
\toprule 
\multirow{2}{*}{Algorithm} & \multirow{2}{*}{Architecture}  & \multirow{2}{*}{Pretraining}  & \multirow{2}{*}{PACS}  & \multirow{2}{*}{VLCS}  & \multirow{2}{*}{OfficeHome}  & \multirow{2}{*}{TerraIn.}  & \multirow{2}{*}{DomainNet}  & \multirow{2}{*}{Avg.} & \multirow{2}{*}{\#Param.}  & Trainable  \\
 & & & & & & & & & & \#Param. \\ 
\midrule 
ERM$^\dagger$ & ViT-B/16 & CLIP$_\textrm{OpenAI}$ & 83.4 \scriptsize{$\pm$ 0.5} & 75.9 \scriptsize{$\pm$ 1.3} & 66.4 \scriptsize{$\pm$ 0.5} & 35.3 \scriptsize{$\pm$ 0.8} & 44.4 \scriptsize{$\pm$ 0.6} & 61.1 & 85.8M & 85.8M \\
MIRO$^\dagger$& ViT-B/16 & CLIP$_\textrm{OpenAI}$ & 95.6 \scriptsize{$\pm$ 0.8} & 82.2 \scriptsize{$\pm$ 0.3} & 82.5 \scriptsize{$\pm$ 0.1} & 54.3 \scriptsize{$\pm$ 0.4} & 54.0 \scriptsize{$\pm$ 0.3} & 73.7 & 172M & 85.8M \\
MIRO$^*$& ViT-B/16 & CLIP$_\textrm{LAION2B}$ & 96.7 \scriptsize{$\pm$ 0.7} & 82.4 \scriptsize{$\pm$ 0.3} & 87.3 \scriptsize{$\pm$ 0.5} & 52.3 \scriptsize{$\pm$ 0.5} & 50.6 \scriptsize{$\pm$ 0.6} & 73.9 & 172M & 85.8M \\
ERM$^\dagger$ & RegNetY-16GF & SWAG$_\textrm{IG3B}$ & 89.6 \scriptsize{$\pm$ 0.4} & 78.6 \scriptsize{$\pm$ 0.3} & 71.9 \scriptsize{$\pm$ 0.6} & 51.4 \scriptsize{$\pm$ 1.8} & 48.5 \scriptsize{$\pm$ 0.6} & 68.0 & 83.6M & 83.6M \\
MIRO$^\dagger$& RegNetY-16GF & SWAG$_\textrm{IG3B}$ & 97.4 \scriptsize{$\pm$ 0.2} & 79.9 \scriptsize{$\pm$ 0.6} & 80.4 \scriptsize{$\pm$ 0.2} & 58.9 \scriptsize{$\pm$ 1.3} & 53.8 \scriptsize{$\pm$ 0.1} & 74.1 & 167.2M & 83.6M \\
ERM+SWAD$^\dagger$& RegNetY-16GF & SWAG$_\textrm{IG3B}$ & 94.7 \scriptsize{$\pm$ 0.2} & 79.7 \scriptsize{$\pm$ 0.2} & 80.0 \scriptsize{$\pm$ 0.1} & 57.9 \scriptsize{$\pm$ 0.7} & 53.6 \scriptsize{$\pm$ 0.6} & 73.2 & 83.6M & 83.6M \\
MIRO+SWAD$^\dagger$& RegNetY-16GF & SWAG$_\textrm{IG3B}$ & 96.8 \scriptsize{$\pm$ 0.2} & 81.7 \scriptsize{$\pm$ 0.1} & 83.3 \scriptsize{$\pm$ 0.1 }& 64.3 \scriptsize{$\pm$ 0.3} & 60.7 \scriptsize{$\pm$ 0.0} & 77.3 & 167.2M & 83.6M \\  
SMA         & RegNetY-16GF & SWAG$_\textrm{IG3B}$ & 95.5 \scriptsize{$\pm$ 0.0} & 80.7 \scriptsize{$\pm$ 0.1} & 82.0 \scriptsize{$\pm$ 0.0} & 59.7 \scriptsize{$\pm$ 0.0} & 60.0 \scriptsize{$\pm$ 0.0} & 75.6 & 83.6M & 83.6M \\
\midrule
\multicolumn{11}{l}{\textit{Methods using additional supervision}} \\
\midrule
\textcolor{gray}{\textit{VL2V-ADiP}} & \textcolor{gray}{\textit{ViT-B/16}} & \textcolor{gray}{\textit{CLIP$_\textrm{OpenAI}$}} & \textcolor{gray}{\textit{94.9}} & \textcolor{gray}{\textit{81.9}} & \textcolor{gray}{\textit{85.7}} & \textcolor{gray}{\textit{55.4}} & \textcolor{gray}{\textit{59.4}} & \textcolor{gray}{\textit{75.5}} & \textcolor{gray}{\textit{235.8M}} & \textcolor{gray}{\textit{83.6M}}\\
\textcolor{gray}{\textit{VL2V-SD}} & \textcolor{gray}{\textit{ViT-B/16}} & \textcolor{gray}{\textit{CLIP$_\textrm{OpenAI}$}} & \textcolor{gray}{\textit{96.7}} & \textcolor{gray}{\textit{83.3}} & \textcolor{gray}{\textit{87.4}} & \textcolor{gray}{\textit{58.5}} & \textcolor{gray}{\textit{62.8}} & \textcolor{gray}{\textit{77.7}} & \textcolor{gray}{\textit{235.8M}} & \textcolor{gray}{\textit{83.6M}} \\
\midrule
\multicolumn{11}{l}{\textit{Methods with Parameter-Efficient Fine-Tuning (PEFT)}} \\
\midrule
\rowcolor{blue!10}
ERM (Baseline) & ViT-B/16 & CLIP$_\textrm{LAION2B}$ & 85.8 \scriptsize{$\pm$ 2.1} & 78.5 \scriptsize{$\pm$ 0.9}& 78.1 \scriptsize{$\pm$ 0.8} & 41.0 \scriptsize{$\pm$ 1.6} & 52.2 \scriptsize{$\pm$0.1} & 67.1 & 85.8M & 85.8M \\
\rowcolor{blue!10}
ERM$_\textrm{Compacter}$ & ViT-B/16 & CLIP$_\textrm{LAION2B}$ & 94.1 \scriptsize{$\pm$ 0.4} & 81.0 \scriptsize{$\pm$ 0.5} & 83.0 \scriptsize{$\pm$ 0.1} & 35.9 \scriptsize{$\pm$0.7} & 56.2 \scriptsize{$\pm$1.2} & 70.0 & 85.9M & 0.10M \\
\rowcolor{blue!10}
ERM$_\textrm{Attention}$ & ViT-B/16 & CLIP$_\textrm{LAION2B}$ & 93.8 \scriptsize{$\pm$ 0.6} & 82.0 \scriptsize{$\pm$ 0.3} & 85.9 \scriptsize{$\pm$ 0.4} & 51.4 \scriptsize{$\pm$0.8} & 57.2 \scriptsize{$\pm$0.1} & 74.1 & 85.8M & 28.4M \\
\rowcolor{blue!10}
ERM$_{\textrm{LoRA},r=2}$ & ViT-B/16 & CLIP$_\textrm{LAION2B}$ & 96.4 \scriptsize{$\pm$ 0.6} & 82.6 \scriptsize{$\pm$ 0.6} & 86.7 \scriptsize{$\pm$ 0.3} & 46.1 \scriptsize{$\pm$1.7} & 61.5 \scriptsize{$\pm$0.1} & 74.7 & 85.9M & 0.11M \\
\rowcolor{blue!10}
ERM$_\textrm{KAdaptaion}$ & ViT-B/16 & CLIP$_\textrm{LAION2B}$ & 97.5 \scriptsize{$\pm$ 0.1} & 83.0 \scriptsize{$\pm$ 0.1} & 90.3 \scriptsize{$\pm$ 0.1} & 51.9 \scriptsize{$\pm$0.5} & \underline{62.7} \scriptsize{$\pm$0.0} & 77.1 & 85.9M & 0.14M \\
\midrule
\multicolumn{11}{l}{\textit{Methods with Mixture-of-Adapter (MoA)}} \\
\midrule
\rowcolor{blue!10}
\rowcolor{blue!10}
ERM$_\textrm{KAdaptaion-MoA}$ & ViT-B/16 & CLIP$_\textrm{OpenAI}$ & 96.2 \scriptsize{$\pm$0.1} & 83.6 \scriptsize{$\pm$0.6} & 84.5 \scriptsize{$\pm$0.0}& 54.3 \scriptsize{$\pm$1.4} & 59.9 \scriptsize{$\pm$0.1} & 75.7 & 87.3M & 1.5M \\
\rowcolor{blue!10}
ERM$_{\textrm{LoRA-MoA}}$ & ViT-B/16 & CLIP$_\textrm{LAION2B}$ & 96.9 \scriptsize{$\pm$ 0.3} & 82.8 \scriptsize{$\pm$ 0.7} & 89.5 \scriptsize{$\pm$ 0.2} & 49.2 \scriptsize{$\pm$ 2.4} & 62.2 \scriptsize{$\pm$ 0.0} & 75.9 & 87.2M & 1.5M \\
\rowcolor{blue!10}
ERM$_\textrm{KAdaptaion-MoA}$ & ViT-B/16 & CLIP$_\textrm{LAION2B}$ & 97.4 \scriptsize{$\pm$ 0.2} & \underline{83.1} \scriptsize{$\pm$ 0.3} & \underline{90.6} \scriptsize{$\pm$ 0.0} & 52.8 \scriptsize{$\pm$ 1.4} & \underline{62.7} \scriptsize{$\pm$ 0.1} & 77.3 & 87.3M & 1.5M \\
\midrule
\multicolumn{11}{l}{\textit{Methods using ensemble}} \\
\midrule
\rowcolor{gray!10}
EoA         & RegNetY-16GF & SWAG$_\textrm{IG3B}$ & 95.8 & 81.1 & 83.9 & \textbf{61.1} & 60.9 & 76.6 & $>$ 500M & - \\  
\rowcolor{gray!10}
SIMPLE      & ModelPool-A & ModelPool-A & 88.6 \scriptsize{$\pm$ 0.4} &  79.9 \scriptsize{$\pm$ 0.5} & 84.6 \scriptsize{$\pm$ 0.5} & 57.6 \scriptsize{$\pm$ 0.8} & 49.2 \scriptsize{$\pm$ 1.1} & 72.0 & $>$ 1,000M & 0.9M \\
\rowcolor{gray!10}
SIMPLE$^+$  & ModelPool-B & ModelPool-B & \textbf{99.0} \scriptsize{$\pm$ 0.1} & 82.7 \scriptsize{$\pm$ 0.4} & 87.7 \scriptsize{$\pm$ 0.4}  & \underline{59.0} \scriptsize{$\pm$ 0.6} & 61.9 \scriptsize{$\pm$ 0.5} & \textbf{78.1} & $>$ 1,000M & 0.9M \\
\rowcolor{blue!10}
ERM$_\textrm{KAdaptaion-MoA-Ensemble}$ & ViT-B/16 & CLIP$_\textrm{LAION2B}$ & \underline{97.6} & \textbf{83.4} & \textbf{90.9} & 54.3 & \textbf{63.1} & \underline{77.9} & 261.3M & - \\
\bottomrule
\end{tabular}
} \vspace{-5pt}
 \caption{
 Quantitative evaluation on domain generalization datasets with MoA methods. The numbers marked with $^\dagger$ are reported results from \cite{cha2022domain}, and those marked with $^*$ are reproduced results conducted by us. OpenAI denotes the private dataset used to train CLIP~\cite{radford2021learning}, and ModelPool-A and B denotes the set of pretrained models used in \cite{li2023simple}. Results with ensemble methods are colored in gray, and results with our method are colored in blue. 
 Additionally, results with additional supervision, such as VL2V-ADiP~\cite{addepalli2024leveraging}, are shown in gray font.
 }\label{tab:moe}\vspace{-10pt}
\end{table*}

\section{The Effectiveness of Parameter-Efficient Fine-Tuning in Domain Generalization}
\label{sec:4}
In this section, we investigate how adapters impact loss landscape flatness by analyzing the maximum Hessian eigenvalue spectra and loss landscapes of various trained models. It is well established that finding flat minima is closely related to the generalization performance and robustness of trained models, as demonstrated by prior studies~\cite{izmailov2018averaging,keskar2016large,foret2020sharpness,wortsman2022model}. Therefore, we compare loss landscapes and maximum Hessian eigenvalues to predict which model is likely to generalize better to unseen domains. We present the results of our analysis in the following sections.

\subsection{Loss Landscapes}
\label{sec:loss_landscape}
As shown in \cite{park2022vision, garipov2018loss, li2018visualizing}, by randomly perturbing the trained weights with random direction vectors, we can obtain variations in the loss value that can be used to predict the model's generalization ability~\cite{foret2020sharpness,cha2021swad}. We evaluate this on PACS~\cite{li2017deeper}, one of the datasets in the DG benchmark~\cite{gulrajani2020search}. As shown in Fig.~\ref{fig:loss_surface}, compared to the case of full fine-tuning (Fig.~\ref{fig:loss_surface_fullft}), where we trained all of the model parameters, the model parameters tuned with LoRA~\cite{hu2021lora,simo2023lora} (Fig.~\ref{fig:loss_surface_lora}) and KAdaptation~\cite{he2022parameter} (Fig.~\ref{fig:loss_surface_ka}) display a flatter loss surface. Additionally, KAdaptation shows a flatter loss surface than LoRA. Further visualizations of the loss landscapes for all test environments in the PACS dataset are provided in the supplemental materials.
\begin{figure}[t]
\centering
    \begin{subfigure}[t]{0.45\linewidth}
        \includegraphics[width=\linewidth]{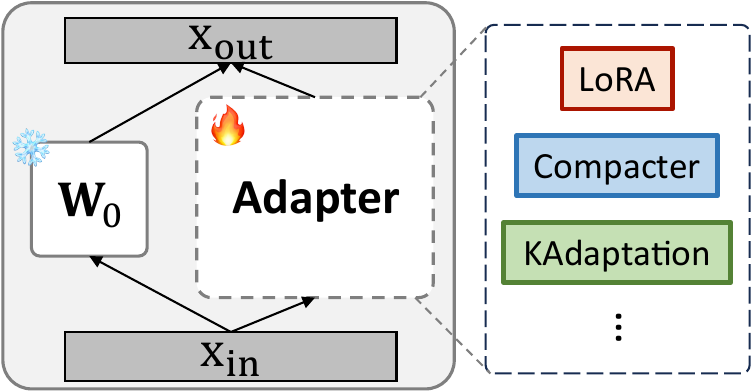}
            \caption{Adapter PEFT methods}
            \label{fig:arch-a}
    \end{subfigure}
        \hspace{0.3em}
    \begin{subfigure}[t]{0.45\linewidth}
            \includegraphics[width=\linewidth]{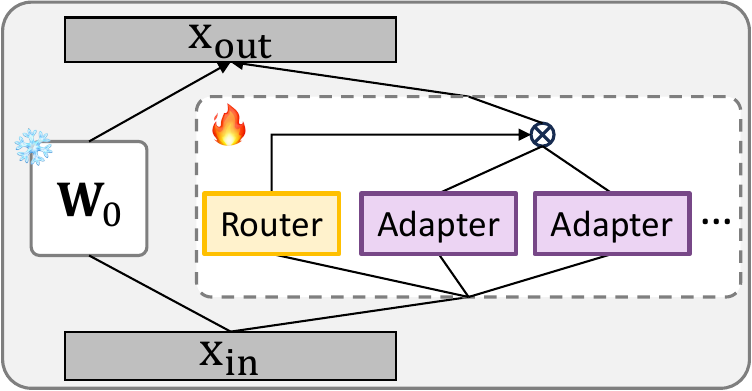}
            \caption{Mixture-of-Adapter method}
            \label{fig:arch-b}
    \end{subfigure}
    \vspace{-5pt}
        \caption{Architecture of the proposed Mixture-of-Adapters (MoA). $\mathbf{W}_0$, $\mathbf{x}_\mathrm{in}$, and $\mathbf{x}_\mathrm{out}$ denotes original pretrained weight, input, and output tokens in multi-head self-attention (MHSA). The \texttt{Adapter} in (a) can refer to any adapter-based PEFT method, such as LoRA, Compacter, or KAdaptation. Additionally, the \texttt{Router} in (b) can be a linear or cosine router, as commonly used in Mixture-of-Expert methods.} \vspace{-10pt}
    \label{fig:arch-moa}
\end{figure}

\subsection{Maximum Hessian Eigenvalue Spectra}
\label{sec:hessian_eigenvalue}
Drawing loss landscapes from the optimal point to random directions sometimes fail to fully represent the shape of loss surfaces due to the high dimensionality of the loss surface~\cite{garipov2018loss, li2018visualizing}. Therefore, following \cite{park2022vision} we calculate the top-5 Hessian eigenvalues and show their spectra. 

Analogous to the loss landscape results in the previous section, we observe the same phenomena with the use of adapters such as LoRA and KAdaptation. As depicted in Fig.~\ref{fig:hessian}, the top-5 Hessian eigenvalues are more concentrated around zero than those of the fully fine-tuned model. This suggests that employing adapters with large pretrained models results in a flatter loss surface around the optimal point. The Hessian matrix serves as an indicator of the curvature of the loss surface. Since the flatness of loss surfaces has a substantial impact on model performance in DG tasks~\cite{cha2021swad}, it can be concluded that adapter fine-tuning with large pretrained models effectively facilitates improved performance in domain generalization.

\subsection{Parameter-Efficient Adapter for Domain Generalization}
Previously, we demonstrated that models trained with PEFT methods can reach a more generalizable optimization point. However, it’s difficult to conclusively determine that a flatter loss surface directly leads to better generalization. Therefore, in the following section, we conduct a comprehensive evaluation of various fine-tuning methods, compare them, and discuss the most effective practical approach while exploring strategies to further enhance performance.

In Table~\ref{tab:peft-compare}, we present results from several studies across five datasets, namely PACS~\cite{li2017deeper}, VLCS~\cite{fang2013unbiased}, OfficeHome~\cite{venkateswara2017deep}, TerraIncognita~\cite{beery2018recognition} and DomainNet~\cite{peng2019moment}. Based on the observations above, we evaluate ERM with various partial fine-tuning and PEFT adapter methods in the standard DG benchmark from \cite{gulrajani2020search}. Some works like MIRO~\cite{cha2022domain} report results of full fine-tuning a CLIP-ViT-B/16 model with an ERM algorithm. However we find that training just a small adapter layer, attention layer, or only bias layer yields a greater performance. When compared to methods that use additional regularization or ensembling, our parameter-efficient training approach with a straightforward ERM algorithm achieves similar or even superior accuracy on the OfficeHome dataset. Moreover, the KAdaptation method delivers the highest average accuracy among all evaluated approaches. These results validate our observation in Sec.~\ref{sec:loss_landscape} and \ref{sec:hessian_eigenvalue}, demonstrating that a proper selection of an adapter method can boost the performance.

\section{Mixture-of-Adapters (MoA)}
\label{sec:moe}

\begin{figure*}[t]
    \centering
    \begin{subfigure}[t]{1.0\textwidth}
    \centering
        \begin{subfigure}[t]{0.195\textwidth}
            \includegraphics[width=\textwidth]{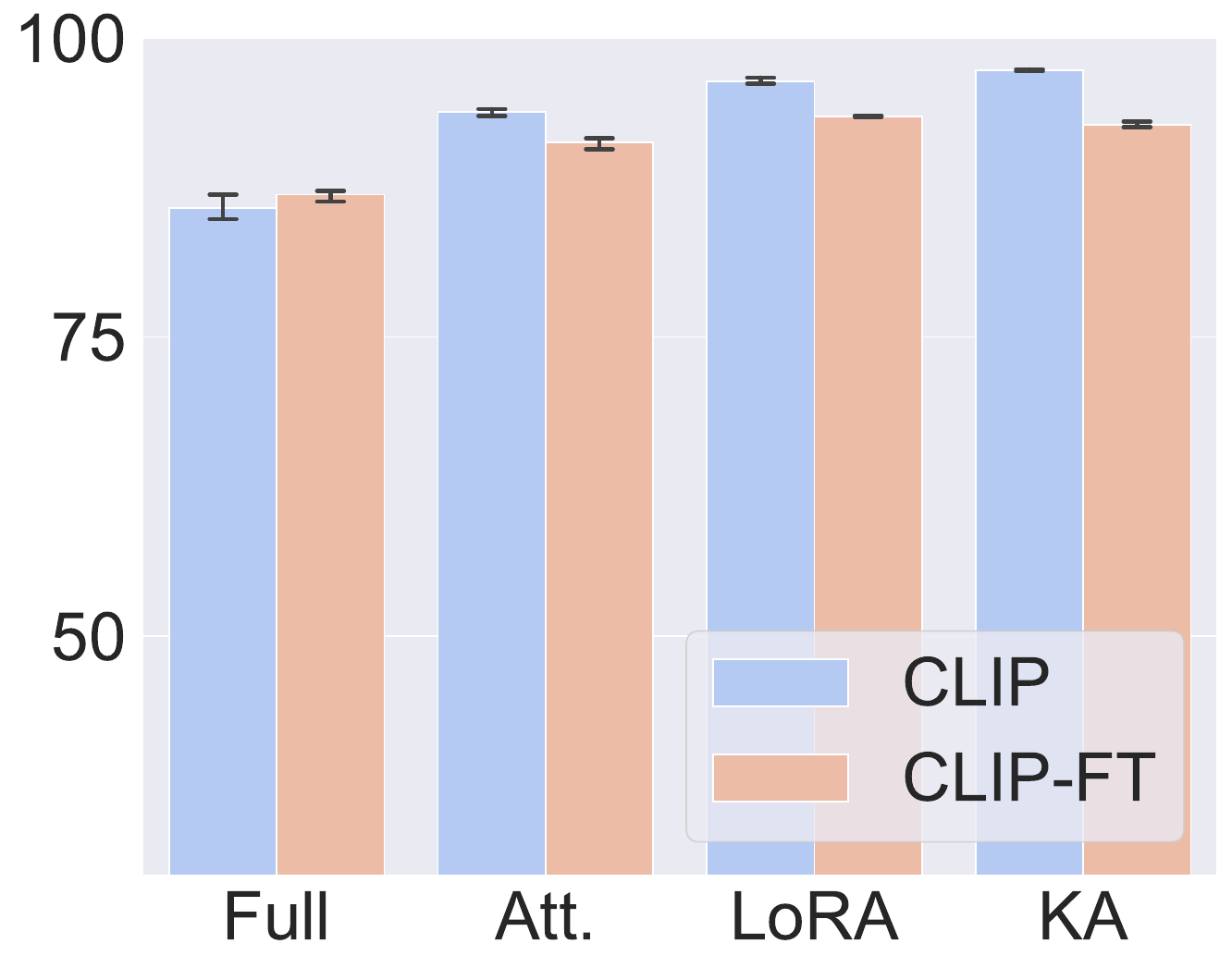}
            \caption{PACS~\cite{li2017deeper}}
        \end{subfigure}
        \begin{subfigure}[t]{0.195\textwidth}
            \includegraphics[width=\textwidth]{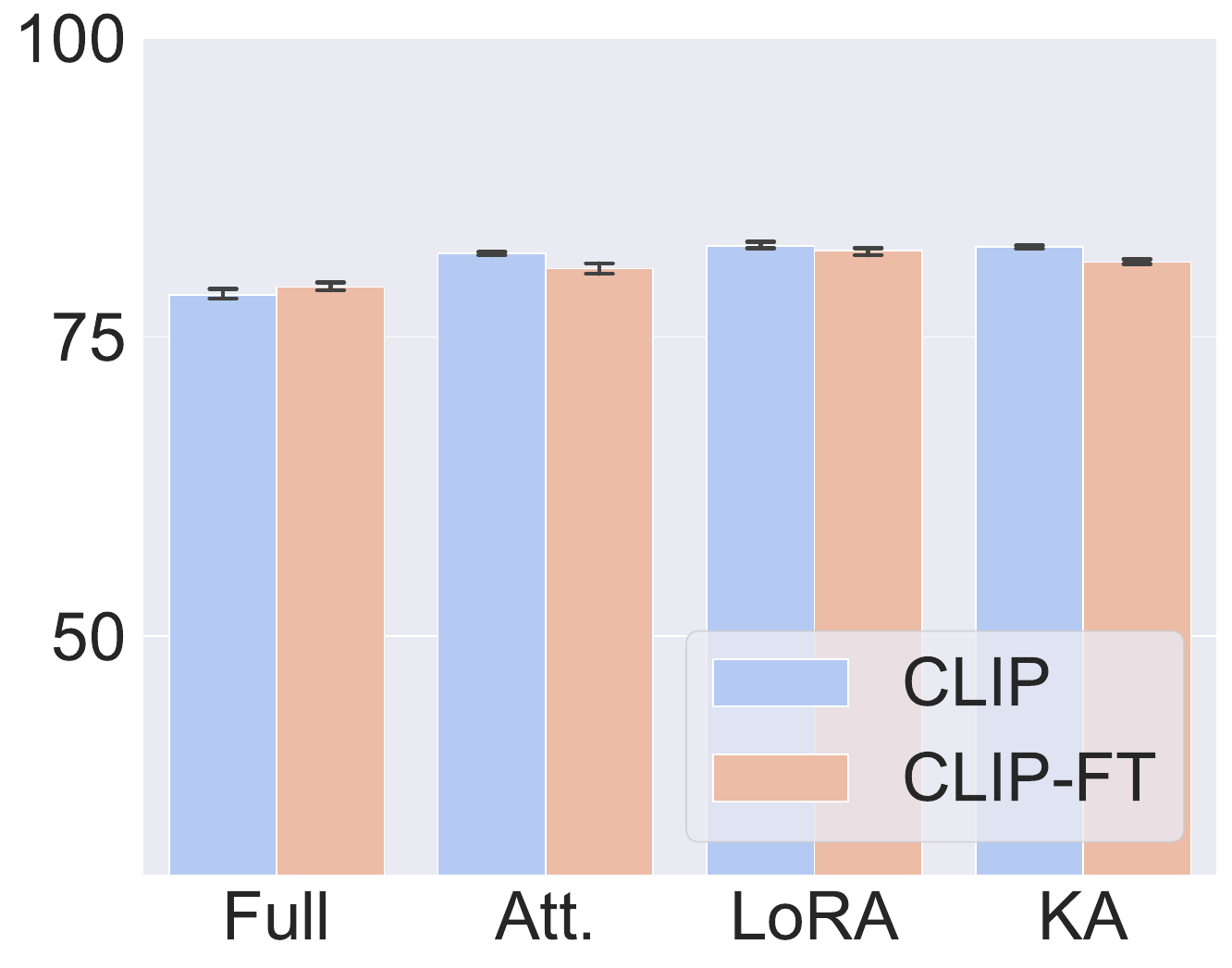}
            \caption{VLCS~\cite{fang2013unbiased}}
        \end{subfigure}
        \begin{subfigure}[t]{0.195\textwidth}
            \includegraphics[width=\textwidth]{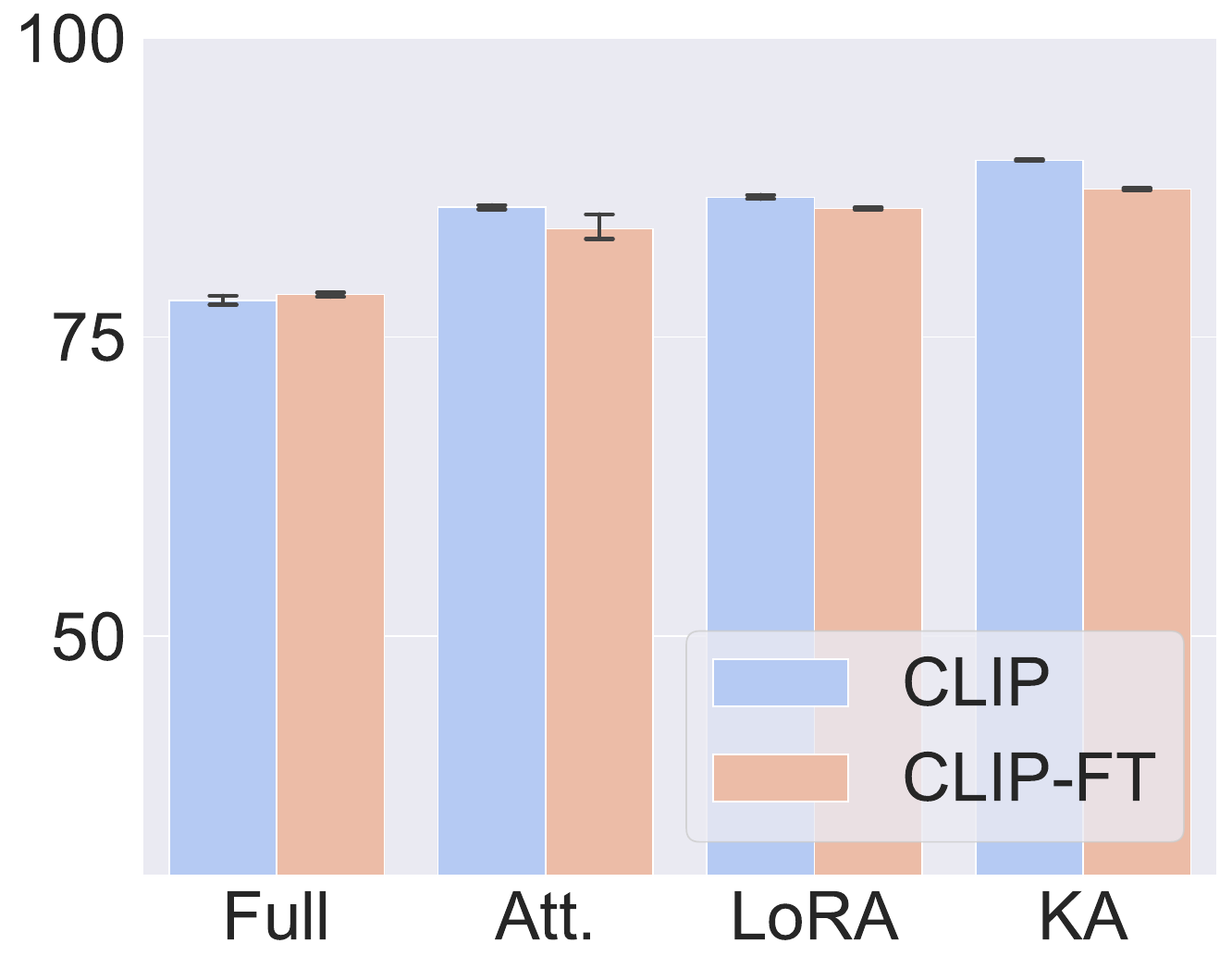}
            \caption{OfficeHome~\cite{venkateswara2017deep}}
        \end{subfigure}
        \begin{subfigure}[t]{0.195\textwidth}
            \includegraphics[width=\textwidth]{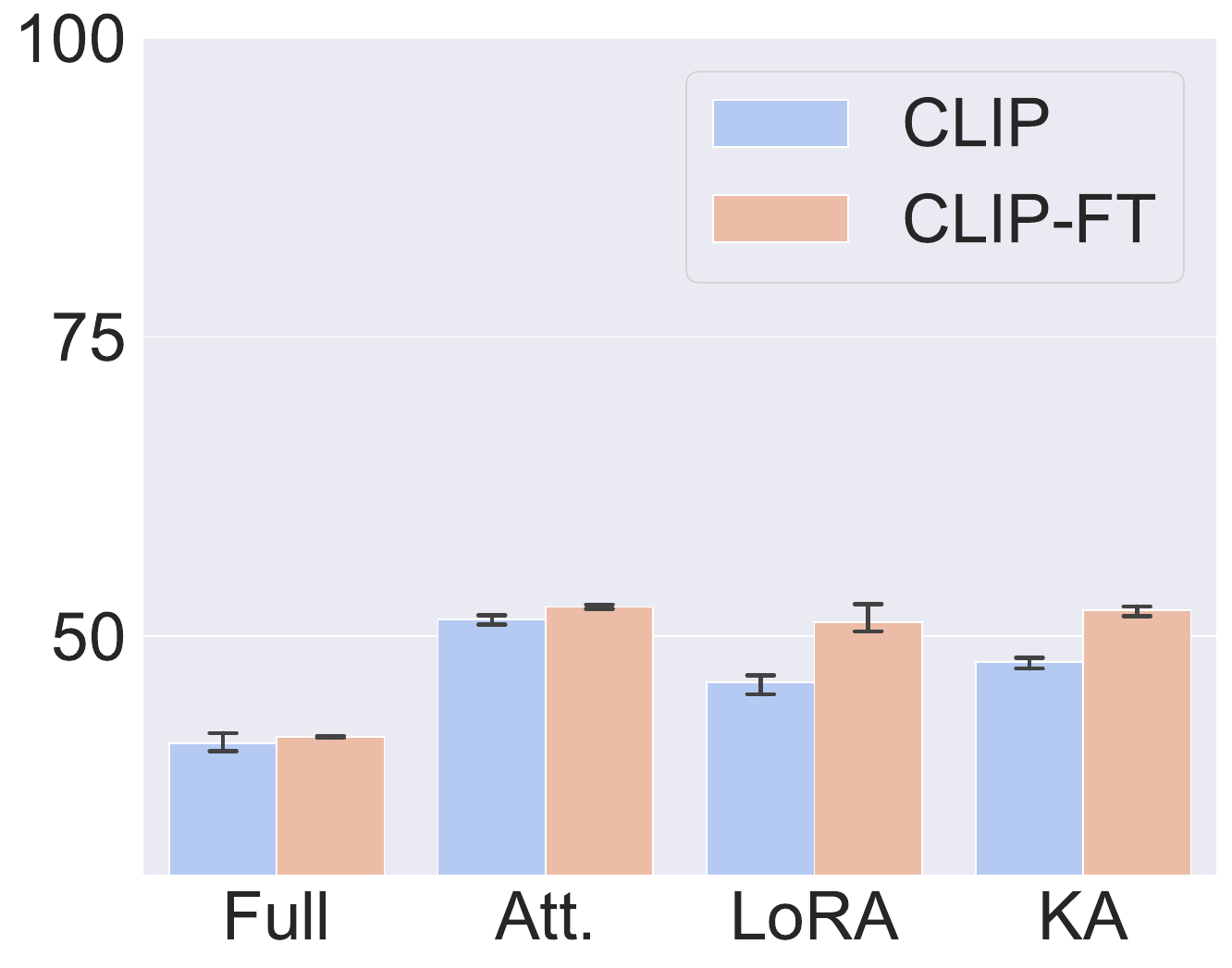}
            \caption{TerraIncognita~\cite{beery2018recognition}}
        \end{subfigure}
        \begin{subfigure}[t]{0.195\textwidth}
            \includegraphics[width=\textwidth]{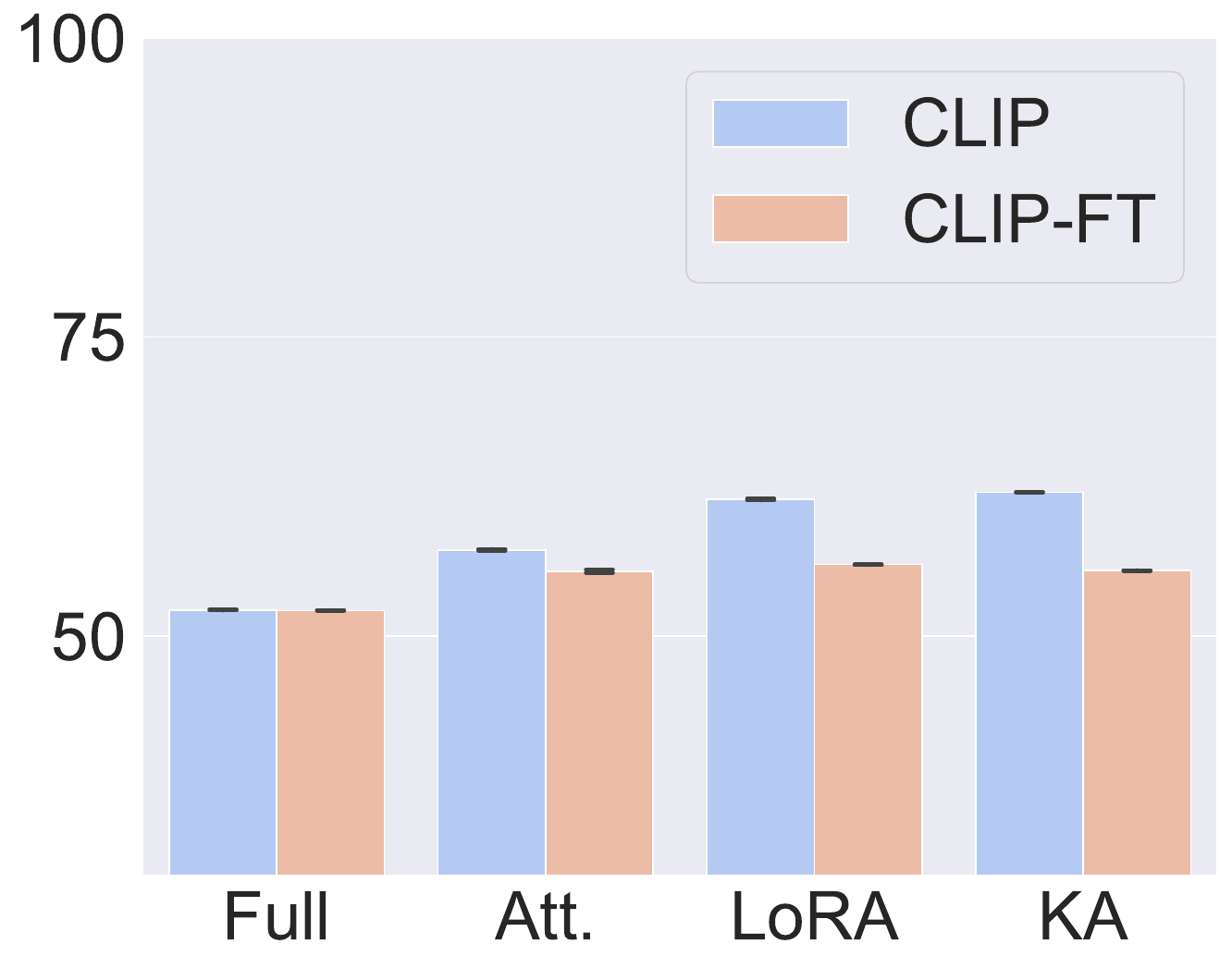}
            \caption{DomainNet~\cite{peng2019moment}}
        \end{subfigure}
    \end{subfigure} \vspace{-5pt}
    \caption{Performance comparison between the original (not fine-tuned) CLIP and the ImageNet fine-tuned CLIP from timm~\cite{rw2019timm} (denoted as CLIP-FT) using different fine-tuning strategies. Full, Att., LoRA, and KA denotes full fine-tuning, attention-only tuning~\cite{touvron2022three}, LoRA, and KAdaptation, respectively.} 
    \label{fig:finetune-compare}\vspace{-10pt}
\end{figure*}

In this section, we demonstrate the effectiveness of our proposed method, mixture-of-adapter (MoA), which tunes large models at low cost while achieving high performance on DG tasks. As highlighted in~\cite{li2022sparse}, MoE architectures utilizing cosine routing are more effective for DG tasks. These architectures also manage various visual attributes, thereby minimizing incorrect token allocation due to intra-domain similarities. We adapt this concept to our adapter-based method, which maintains computational efficiency while delivering improved results in domain generalization.
As outlined in Sec.~\ref{sec:intro}, we utilize adapters to handle various amounts of distribution shift among the domains. Specifically, adapters with varying capacities at each MoA layer are used to select the top-k outputs, which are then averaged and combined with the outputs from the original layer. We control the capacity of each adapter by adjusting its inner rank. 
The MoE layer utilizing the cosine router $\mathrm{G}(\mathbf{x})$ and adapter $\mathrm{A}_{r_i}$ with inner rank $r_i$ can be denoted as:
\begin{equation}\begin{split}
&f_\mathrm{MoE}(\mathbf{x})=\sum_{i=1}^{N}\mathrm{G}(\mathbf{x})_{i}\mathrm{A}_{r_i}(\mathbf{x})\\
&=\sum_{i=1}^{N}\mathrm{TOP}_k\left(\texttt{Softmax}\left(\frac{\mathbf{E}^\mathrm{T}\mathbf{W}\mathbf{x}}{{\tau}\|\mathbf{W}\mathbf{x}\| \|\mathbf{E}\|}\right)\right)\mathrm{A}_{r_i}(\mathbf{x})
\end{split}\end{equation} 
where $\mathrm{A}_{r_i}(\mathbf{x})$ is the output of an adapter with a different inner rank $r_i$, $\mathbf{E}\in\mathbb{R}^{e_i \times N}$ is a learned embedding   $W\in\mathbb{R}^{e_i \times r}$ is the projection matrix to the hypersphere and $\tau$ is a learnable temperature term of the softmax layer. Specifically, we choose top-k weights according to the cosine similarity between the projected feature $\mathbf{Wx}\in\mathbb{R}^{e_i}$ and learned embedding $\mathbf{E}$. Then the output of adapters with different ranks are combined using these weights.
We integrate adapters into the large pretrained model by attaching them to the attention submodules. In line with traditional MoE approaches, we use routers to allocate tokens to expert adapters. Adapters can take various forms, including LoRA~\cite{hu2021lora}, Compacter~\cite{karimi2021compacter} and KAdaptation~\cite{he2022parameter}, as visualized in Fig.\ref{fig:arch-moa}.
In summary, the analysis of the loss landscape through the maximum Hessian eigenvalue spectra in Sec.~\ref{sec:4} allowed us to evaluate the model's generalization and robustness. Furthermore, using a mixture of adapters in the PEFT framework enabled cost-effective fine-tuning while achieving strong domain generalization.

\section{Experimental Results}
\subsection{Experimental Setting}
\paragraph{Experimental details.}
We use the standard benchmark DomainBed~\cite{gulrajani2020search} for training and evaluating the performance on the DG task. Following this, we use fixed hyperparameters within the same backbone model for all the experiments. We train five types of models which are full fine-tuning, attention tuning, LoRA, KAdaptation, and Compacter. 
We employ CLIP trained ViT~\cite{dosovitskiy2020image} as our initialization model because CLIP carries strong zero-shot ability which can be used to obtain favorable generalization performance. 
In detail we use OpenCLIP~\cite{ilharco_gabriel_2021_5143773}, an open-source re-implementation of CLIP model with LAION-2B dataset~\cite{schuhmann2022laion}, for all our experiments. 

\vspace{-10pt}
\paragraph{Implementation details.}  
\label{sec:impl-details}
Pretrained vision transformer models are sourced from the timm~\cite{rw2019timm} library. We select the inner rank for all adapters to ensure similar trainable parameters (0.1M). LoRA and KAdaptation weights are applied to the QKV projection and projection weight matrices in the self-attention layer, while Compacter adapters are attached after the MLP layer in each block. 
Additional details are included in the supplemental material.

\subsection{Results for Mixture-of-Adapter}
As described in Sec.~\ref{sec:moe}, we implement MoA with LoRA and KAdaptation, with benchmark results presented in Table~\ref{tab:moe}. The application of PEFT alone resulted in performance improvements, and integrating MoA consistently boosted results across all datasets. For instance, LoRA-MoA improved average performance by 1.2pp compared to using only LoRA. Similarly, KAdaptation-MoA achieved state-of-the-art results on VLCS, OfficeHome, and DomainNet, exceeding the original KAdaptation by 0.2pp on average, validating our loss surface analysis in Fig.~\ref{fig:loss_surface}.
Following \cite{arpit2022ensemble}, ensembling three models trained with different seeds further improved performance by 0.6pp on average, achieving the best results in VLCS, OfficeHome, and DomainNet. Although SIMPLE~\cite{li2023simple} achieved an average of 78.1, which is SOTA, it relies on over 100 pretrained models, incurring significant computational costs. In contrast, our method reaches comparable performance using only three models. Additionally, while VL2V-SD and VL2V-ADiP~\cite{addepalli2024leveraging} leveraged CLIP’s text encoder and embeddings to achieve SOTA results, our model achieved similar performance without relying on these components.

\subsubsection{Analysis of the router and experts}
\begin{figure*}[t]
    \centering
    \begin{subfigure}[t]{0.49\textwidth}
        \begin{subfigure}[t]{0.455\textwidth}
            \includegraphics[width=\textwidth]{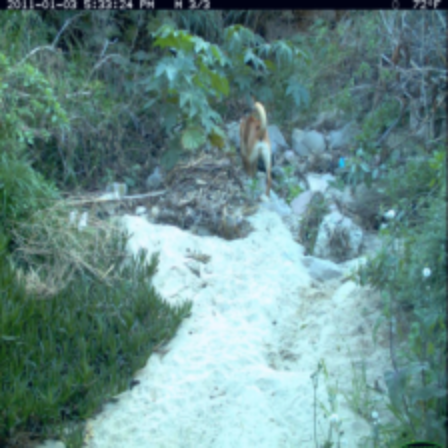}
        \end{subfigure} 
        \begin{subfigure}[t]{0.49\textwidth}
            \includegraphics[width=\textwidth]{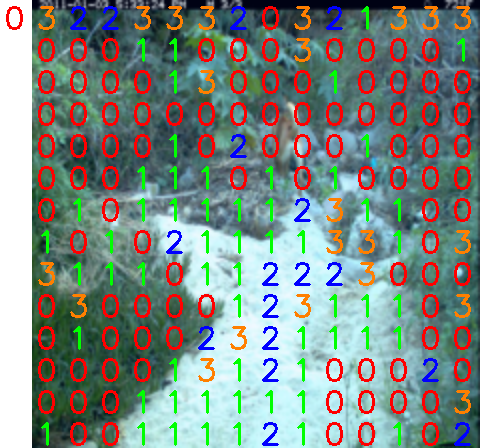}
        \end{subfigure} \hfill
        \\
        \begin{subfigure}[t]{0.455\textwidth}
            \includegraphics[width=\textwidth]{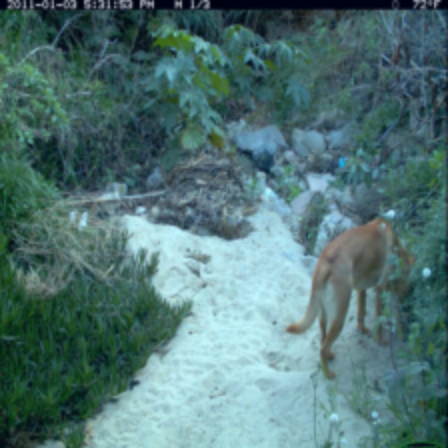}
        \end{subfigure}
        \begin{subfigure}[t]{0.49\textwidth}
            \includegraphics[width=\textwidth]{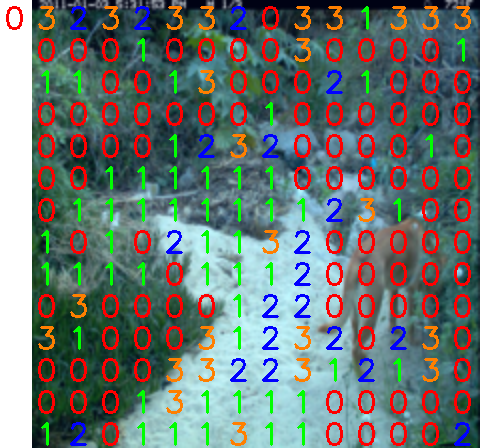}
        \end{subfigure}\hfill
    \caption{Location 100 (class dog)}
    \label{fig:router-terra-a}
    \end{subfigure}
    \centering
    \begin{subfigure}[t]{0.49\textwidth}
        \begin{subfigure}[t]{0.455\textwidth}
            \includegraphics[width=\textwidth]{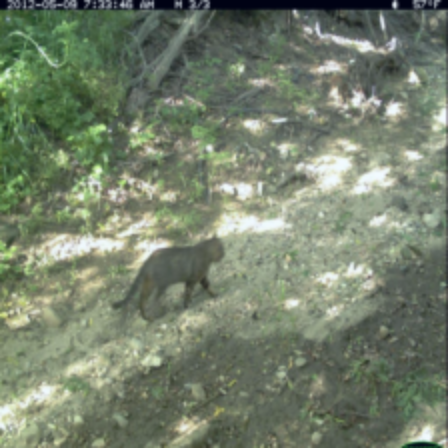}
        \end{subfigure}
        \begin{subfigure}[t]{0.49\textwidth}
            \includegraphics[width=\textwidth]{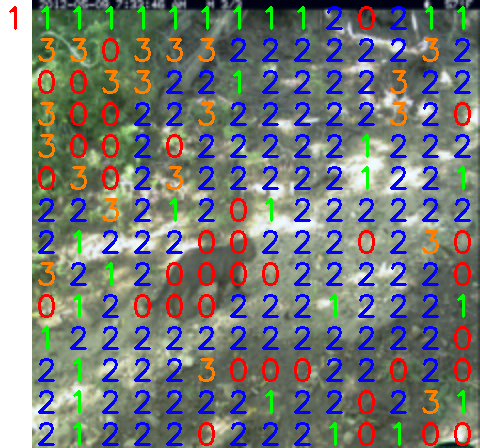}
        \end{subfigure} \hfill
        \\
        \begin{subfigure}[t]{0.455\textwidth}
            \includegraphics[width=\textwidth]{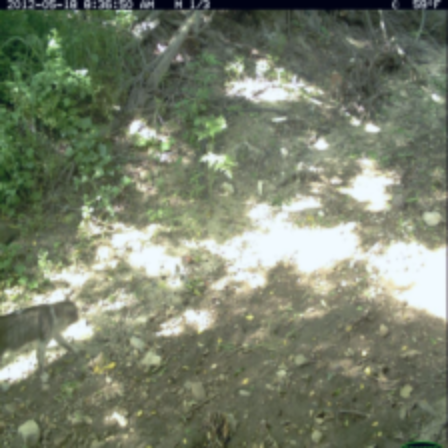}
        \end{subfigure}
        \begin{subfigure}[t]{0.49\textwidth}
            \includegraphics[width=\textwidth]{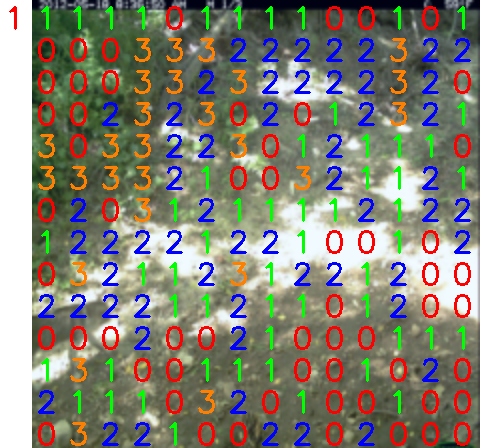}
        \end{subfigure}\hfill
    \caption{Location 38 (class cat)}
    \label{fig:router-terra-b}
    \end{subfigure} 
    \vspace{-5pt} \\
    \caption{Visualizations of routed indices for each patch in the TerraIncognita~\cite{beery2018recognition} dataset. The left column shows the original image, and the right column shows the routed result for each patch. The upper and lower images were captured at the same location but at different times, therefore they share the same background but feature different objects (a dog and a cat) in terms of shape and location.}
    \label{fig:router-analysis}
\end{figure*}

To understand why using multiple adapters with a router improves performance, we conduct experiments to see how the router assigns tokens to each expert and what each expert learns. To investigate the role of an adapter, we monitor the routing path of each token and identify the expert to which it is directed. This visualization is performed on TerraIncognita dataset, one of the hardest datasets in the DG setting. In Fig.~\ref{fig:router-analysis}, the image is divided into patches and each patch is assigned a number, where the number corresponds to the routed expert. We also reveal that the router tends to cluster tokens in areas with semantic information (e.g., object foreground or object outlines). As an example, consider the dog depicted in Fig.~\ref{fig:router-terra-a} and the cat in Fig.~\ref{fig:router-terra-b}, which exhibit varying positions between the first and second rows. 
Images sharing the same location tend to have consistent backgrounds, leading to similar expert routing tendencies for background tokens. However, objects and their positions can differ, causing object-related tokens to be routed to different experts.
For example, in Fig.~\ref{fig:router-terra-a} (a), the upper image shows a dog located almost at the center, while in the bottom image, the dog has moved to the lower-right. As the dog’s location changes, the assigned expert (number 0) also varies to match the new position.
A similar phenomenon is observed in Fig.~\ref{fig:router-terra-b} (b). In the upper image, the cat is almost at the center, with expert 0 assigned to the corresponding patch token. In the bottom image, the cat is in the lower-left, and the token location assigned to expert 0 shifts accordingly.
In conclusion, routing numerous tokens to their respective adapters significantly enhances the model’s ability to capture semantic information in images, enabling it to effectively handle challenging distribution shift scenarios in DG tasks. 
Additional analyses and visualizations of routed patches are provided in the supplemental materials.

\begin{table*}[t]
\centering
\resizebox{\textwidth}{!}{
\begin{tabular}{l|c|ccccc|c|cc}
\toprule 
\multirow{2}{*}{Adapter} & \multirow{2}{*}{Changed component} & \multirow{2}{*}{PACS}  & \multirow{2}{*}{VLCS}  & \multirow{2}{*}{OfficeHome}  & \multirow{2}{*}{TerraIn.}  & \multirow{2}{*}{DomainNet}  & \multirow{2}{*}{Avg.} & \multirow{2}{*}{\#Param.} & Trainable  \\
 & & & & & & & & & \#Param. \\ 
\midrule 
\rowcolor{blue!10} %
\multirow{4}{*}{\cellcolor{white} KAdaptation-MoA} & Original & \textbf{97.5} & \textbf{82.8} & \textbf{90.6} & 53.1 & \textbf{62.6} & 77.3 & 87.3M & 1.5M \\ \cmidrule{2-10}
                                & w/o $\mathcal{L}_\mathrm{aux}$ & 97.3 & 82.6 & 90.5 & \textbf{54.0} & 62.6 & 77.4 & 87.3M & 1.5M  \\
                                & \texttt{Cosine} $\rightarrow$ \texttt{Linear} & 97.5 & 82.3 & 90.3 & 51.5 & 62.6 & 76.9 & 86.1M & 0.33M \\
                                & \texttt{Every} $\rightarrow$ \texttt{Last} & 97.2 & 82.2 & 90.2 & 47.5 & 62.4 & 75.9 & 86.3M & 0.51M \\
\bottomrule
\end{tabular}
} \vspace{-5pt}
\caption{Performance comparison on different mixture-of-adapter settings. We evaluate the impact of removing each component from our best-performing configuration (highlighted in blue). `w/o $\mathcal{L}_{\mathrm{aux}}$' indicates results without the auxiliary loss from \cite{li2022sparse}, `\texttt{Cosine}$\rightarrow$\texttt{Linear}' indicates replacing the \texttt{Cosine} router with a \texttt{Linear} router, and `\texttt{Every}$\rightarrow$\texttt{Last}' indicates changing adapter placement from every two layers to the last two layers.
}
\label{tab:abl-router}
\vspace{-10pt}
\end{table*}

\subsection{Ablation Study}

\subsubsection{Effect of fine-tuning in pretrained models}
We observe that fine-tuning with a smaller dataset degrades performance. Specifically, LAION-2B~\cite{schuhmann2022laion} pretrained CLIP-ViT model almost consistently outperforms the LAION-2B pretrained, ImageNet~\cite{deng2009imagenet} fine-tuned CLIP-ViT model across all the adapter methods in terms of accuracy except for TerraIncognita dataset, as illustrated in Fig.~\ref{fig:finetune-compare}.
Additionally, adapter and partial fine-tuning methods consistently outperform full fine-tuning in both CLIP and CLIP-FT, highlighting the loss of generalization performance during fine-tuning. This can be attributed to larger models’ superior generalization capabilities and adapters’ ability to retain pretrained knowledge.
These findings align with prior work, such as \cite{kumar2022fine}, which shows that fully fine-tuning large models can degrade learned representations.

\subsubsection{Ablation study on routing strategy}
We conduct an ablation study on three components proposed in \cite{li2022sparse}, which are cosine router, auxiliary loss $\mathcal{L}_\mathrm{aux}$, and the layer selection of the MoE layer. \cite{li2022sparse} demonstrates that employing a cosine router in their Generalizable Mixture-of-Experts (GMoE) architecture yields better performance when contrasted with employing a linear router. Our approach deviates from theirs, as each of our expert adapters exhibit distinct capacities, and our MoA is attached to an attention layer. Thus we test both the linear and cosine routers. The auxiliary loss ($\mathcal{L}_\mathrm{aux}$) balances the amount of token allocation to each experts, and the layer configuration called `Every 2' or `Last 2' in their work denotes the attachment of MoE layer with every two layers or only on the last two layers. Therefore we also conduct experiments on these settings and report the whole results in Table~\ref{tab:abl-router}. %
In our setting, employing a cosine router showed a consistent performance increase for all domains. Also attaching an adapter with every two layers brought better performance than using MoA only on the last two layers. While there was a minimal difference on applying $\mathcal{L}_\mathrm{aux}$ (within the error margin), using the auxiliary loss yielded more better results for the four datasets (PACS, VLCS, OfficeHome, DomainNet). Thus we decided to incorporate the auxiliary loss $\mathcal{L}_\mathrm{aux}$ when conducting our main experiment. Additionally, we provide more experiments regarding the effect of $\mathcal{L}_\mathrm{aux}$ on token distribution in the supplemental material.

\vspace{-10pt}
\paragraph{Ablation study about the number of experts}
We conduct experiments to see how performance on the PACS dataset changes with the number of experts, as shown in Table~\ref{tab:num_experts}. Using four experts as the baseline, we see a small performance drop of 0.14 pp when the number of experts is reduced to two, and a drop of 0.12 pp with three experts. Conversely, increasing the number of experts leads to performance gains. Specifically, using five experts lead to a performance gain of 0.04 pp, and using six experts result in a 0.46 pp improvement over the baseline. These results show that our method can be easily scaled to fit the target dataset by adjusting the number of experts.
\begin{table}[t]
    \centering
    \small
    \begin{tabular}{l|ccccc}\toprule
\# of experts & 2     & 3  & 4    & 5     & 6 \\ \midrule
   PACS  & 97.26 & 97.28 & 97.40 & 97.44 & \textbf{97.86} \\ \bottomrule
    \end{tabular}\vspace{-5pt}
    \caption{Performance changes on the PACS dataset based on the number of experts}
    \label{tab:num_experts}
    \vspace{-10pt}
\end{table}

\section{Conclusion}
We have shown that using only parameter-efficient fine-tuning can outperform or be competitive with previous state-of-the-art domain generalization algorithms. We also propose integrating extra adapters with learnable routers to handle various distribution shifts, achieving state-of-the-art results without ensembling and competitive results when ensembling is used.
This work highlights the effectiveness of parameter-efficient fine-tuning and large models for DG, and we hope these findings inspire future research, particularly in robustly fine-tuning large pretrained models.

\noindent\textbf{Acknowledgements.}
This research was supported by Institute of Information \& communications
Technology Planning \& Evaluation (IITP) grant funded by the Korea government (MSIT) (RS-2019II190075, RS-2024-00509279, RS-2020-II201819, RS-2024-00398115, Research on the reliability
and coherence of outcomes produced by Generative AI) and the Culture, Sports, and Tourism R\&D
Program through the Korea Creative Content Agency grant funded by the Ministry of Culture, Sports
and Tourism (RS-2024-00348469, RS-2023-00266509), and National Research Foundation of Korea
(RS-2024-00346597).

\newpage

{\small
\bibliographystyle{ieee_fullname}
\bibliography{egbib}
}
\clearpage
{
\onecolumn

\begin{center}
\Large
\textbf{- Supplementary Materials - \\Domain Generalization using Large Pretrained Models with Mixture-of-Adapters}
\end{center}
\vspace{20pt}
}
\begin{multicols}{2}

\setcounter{section}{0}

In this supplemental material, we provide additional analysis results and visualizations. We also include the code needed to reproduce our experimental results.

\section{Additional implementation details}
\begin{table}[H]
\small
\centering
\resizebox{\linewidth}{!}{

\begin{tabular}{l|cc|cc}
\toprule
{Hyperparameter} & LoRA & KAdaptation & LoRA-MoA & KMoA \\
\midrule
\# of Experts & \multicolumn{2}{c|}{N/A} & $4$ & $4$ \\
Scale of $\mathcal{L}_\mathrm{aux}$ & \multicolumn{2}{c|}{N/A} & $0.01$ & $0.01$ \\
Router & \multicolumn{2}{c|}{N/A} & \multicolumn{2}{c}{Cosine} \\
Router Top-k & \multicolumn{2}{c|}{N/A} & \multicolumn{2}{c}{Top-1} \\
Rank of adapter ($r_i$) & $2$ & $1$ & \multicolumn{2}{c}{$[1, 2, 4, 8]$} \\
\# of Kroneker products ($t$) & N/A & 64 & N/A & 64 \\
Batch size & \multicolumn{4}{c}{$160$ (DomainNet), $96$ (Otherwise)}  \\
Learning rate & \multicolumn{4}{c}{$5e-5$}  \\
Optimizer & \multicolumn{4}{c}{Adam} \\
\bottomrule
\end{tabular}
}
\caption{List of hyperparameters used in experiments on domain generalization benchmarks.}\label{table:hyperparameter} \vspace{-10pt}
\end{table}

All adapters, except LoRA, are implemented from their official repositories; LoRA is implemented using an unofficial version~\cite{simo2023lora}.
By following the previous experimental settings, Adam~\cite{kingma2014adam} optimizer is used for model optimization along with a learning rate of $5e-5$. A batch size of 32 per domain is used for the ViT-Base model. We run 15,000 iterations on DomainNet and 5,000 for others, and evaluate at every 500 iteration steps for DomainNet, 200 steps for others. We perform all experiment on one machine with 8 NVIDIA RTX3090 GPUs.

\vspace{-10pt}
\paragraph{Evaluation protocols and datasets.}
For a fair comparison, we employ DomainBed evaluation protocols~\cite{cha2021swad, gulrajani2020search}. The following five benchmark datasets: PACS~\cite{li2017deeper}, VLCS~\cite{fang2013unbiased}, OfficeHome~\cite{venkateswara2017deep}, TerraIncognita~\cite{beery2018recognition}, and DomainNet~\cite{peng2019moment}. Using a \textit{leave-one-out cross-validation}, all performance scores are evaluated by averaging all the cases that use a single domain as the target domain and the others as the source domains. Experiment is repeated three times and 20\% percent of source domain data is left out for validation purposes. Lastly model selection (training-domain validation) and hyperparameter search follow DomainBed procedures. We perform three runs with different random seeds for each setting and report their mean and standard deviation to show the training randomness. In ablation studies, we keep all the random seeds fixed and conduct the experiment. 
\begin{table}[H]
\footnotesize
\centering
\begin{tabular}{c|c|cccc|c}
    \toprule
    \multirow{2}{*}{Test Env.} & \multirow{2}{*}{$\mathcal{L}_\mathrm{aux}$} & \multicolumn{4}{c|}{Expert} & \multirow{2}{*}{Std} \\
     &  & 0 & 1 & 2 & 3 & \\ \midrule
    \multirow{2}{*}{Art} & \xmark & 0.03 & 0.14 & 0.23 & 0.60  & 0.213\\
                    & \cmark & 0.22 & 0.26 & 0.32 & 0.21  & \textbf{0.044} \\ \midrule
    \multirow{2}{*}{Cartoon} & \xmark & 0.02 & 0.19 & 0.07 & 0.72  & 0.275\\
     & \cmark & 0.23 & 0.18 & 0.25 & 0.35  & \textbf{0.062}\\ \midrule
    \multirow{2}{*}{Photo} & \xmark & 0.07 & 0.17 & 0.56 & 0.21  & 0.184 \\
     & \cmark & 0.23 & 0.21 & 0.29 & 0.26  & \textbf{0.030}\\ \midrule
    \multirow{2}{*}{Sketch} & \xmark & 0.10 & 0.17 & 0.16 & 0.57  & 0.185 \\
     & \cmark & 0.32 & 0.31 & 0.17 & 0.20  & \textbf{0.065}\\ 
    \bottomrule
\end{tabular}
\vspace{-5pt}
\caption{Analysis about the effectiveness of auxiliary loss on PACS dataset. Each number represents the relative allocation ratio, calculated by counting the number of tokens routed to each expert and dividing by the total number of tokens.} \vspace{-10pt}\label{table:laux}
\end{table}

\section{Additional analysis}
In this section, we present an additional analysis of routed tokens, loss landscapes, and maximum Hessian eigenvalue spectra.

\subsection{Comparisons of loss landscape visualizations}
\begin{figure*}[!]
    \centering
    \begin{subfigure}[t]{1.0\textwidth}
        \begin{subfigure}[t]{0.24\textwidth}
            \includegraphics[width=\textwidth]{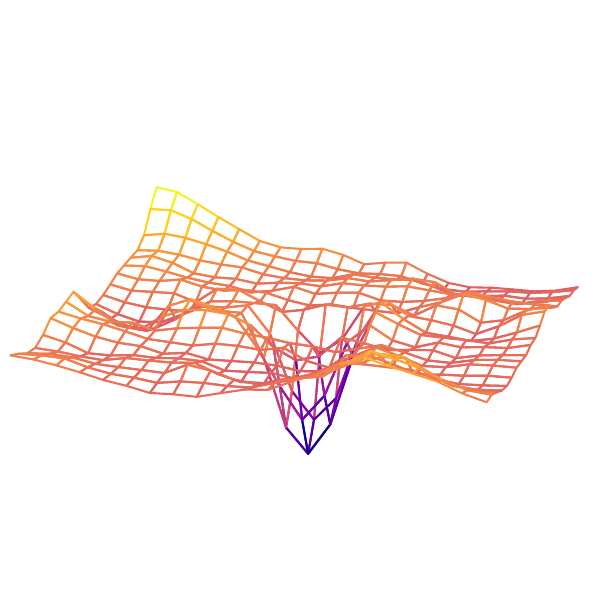}
            \caption*{TE0}
        \end{subfigure}
        \begin{subfigure}[t]{0.24\textwidth}
            \includegraphics[width=\textwidth]{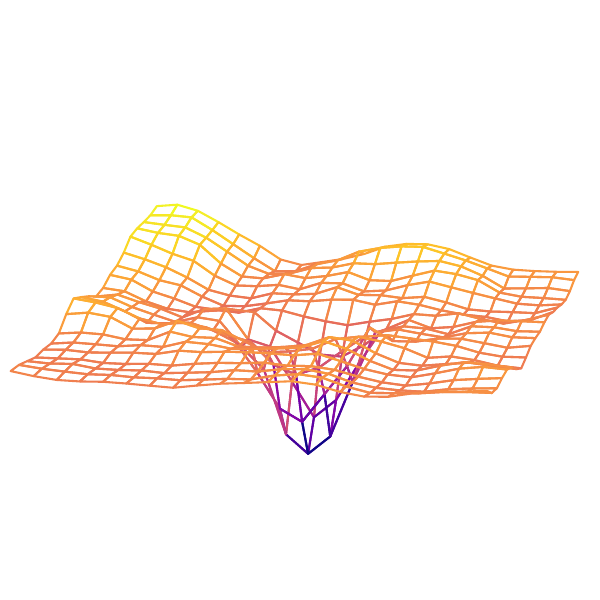}
            \caption*{TE1}
        \end{subfigure}
        \begin{subfigure}[t]{0.24\textwidth}
            \includegraphics[width=\textwidth]{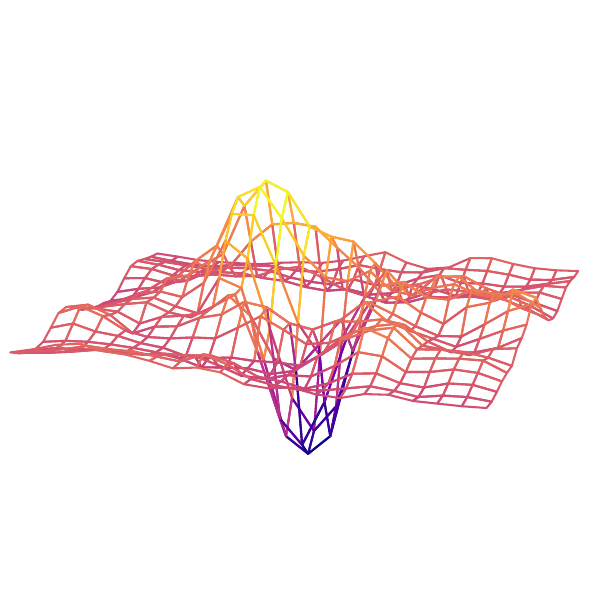}
            \caption*{TE2}
        \end{subfigure}
        \begin{subfigure}[t]{0.24\textwidth}
            \includegraphics[width=\textwidth]{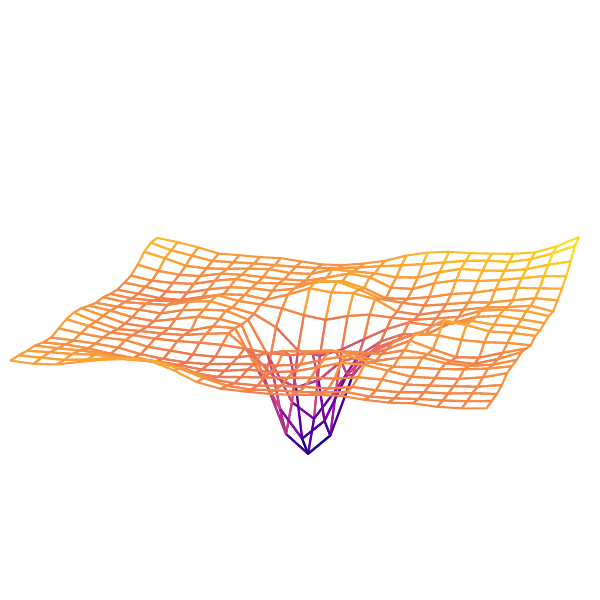}
            \caption*{TE3}
        \end{subfigure}
        \caption{Full fine-tuning}
    \end{subfigure} \\
    \begin{subfigure}[t]{1.0\textwidth}
        \begin{subfigure}[t]{0.24\textwidth}
            \includegraphics[width=\textwidth]{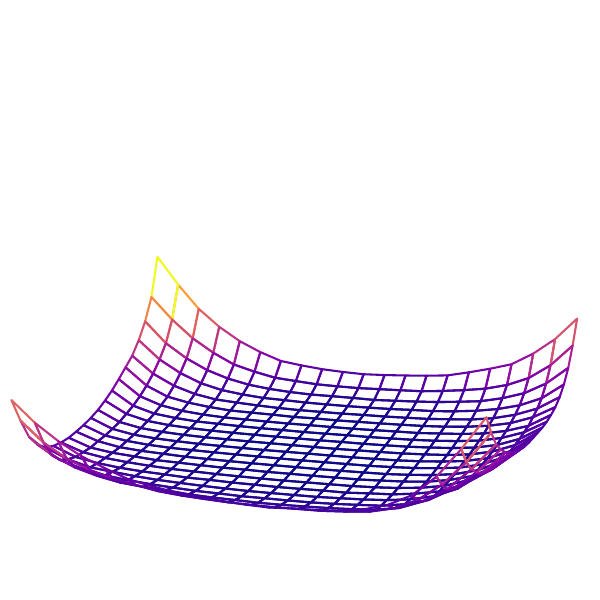}
            \caption*{TE0}
        \end{subfigure}
        \begin{subfigure}[t]{0.24\textwidth}
            \includegraphics[width=\textwidth]{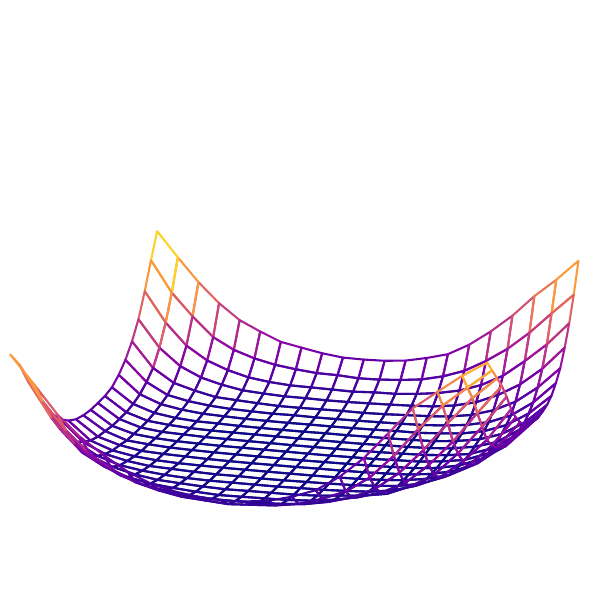}
            \caption*{TE1}
        \end{subfigure}
        \begin{subfigure}[t]{0.24\textwidth}
            \includegraphics[width=\textwidth]{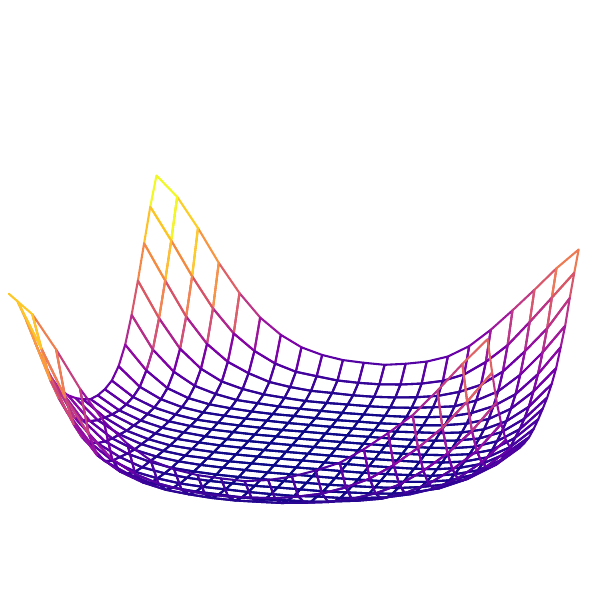}
            \caption*{TE2}
        \end{subfigure}
        \begin{subfigure}[t]{0.24\textwidth}
            \includegraphics[width=\textwidth]{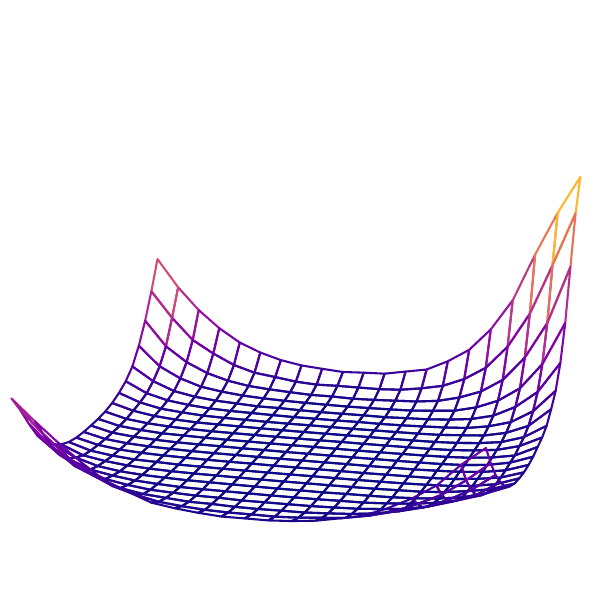}
            \caption*{TE3}
        \end{subfigure}
        \caption{LoRA~\cite{hu2021lora,simo2023lora}}
    \end{subfigure} \\
    \begin{subfigure}[t]{1.0\textwidth}
        \begin{subfigure}[t]{0.24\textwidth}
            \includegraphics[width=\textwidth]{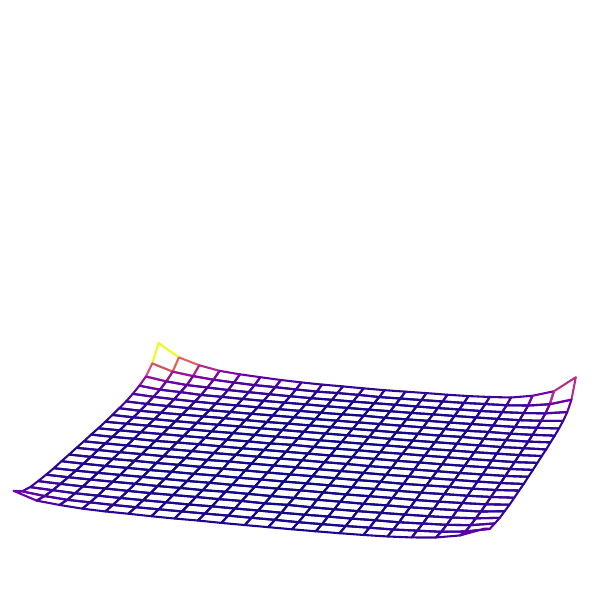}
            \caption*{TE0}
        \end{subfigure}
        \begin{subfigure}[t]{0.24\textwidth}
            \includegraphics[width=\textwidth]{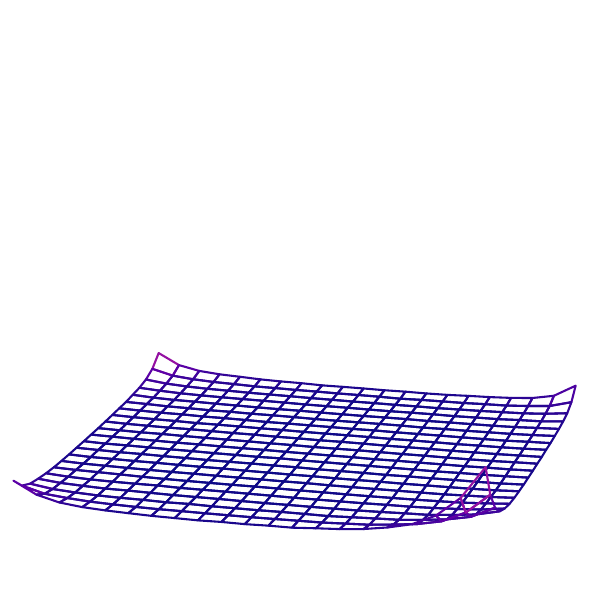}
            \caption*{TE1}
        \end{subfigure}
        \begin{subfigure}[t]{0.24\textwidth}
            \includegraphics[width=\textwidth]{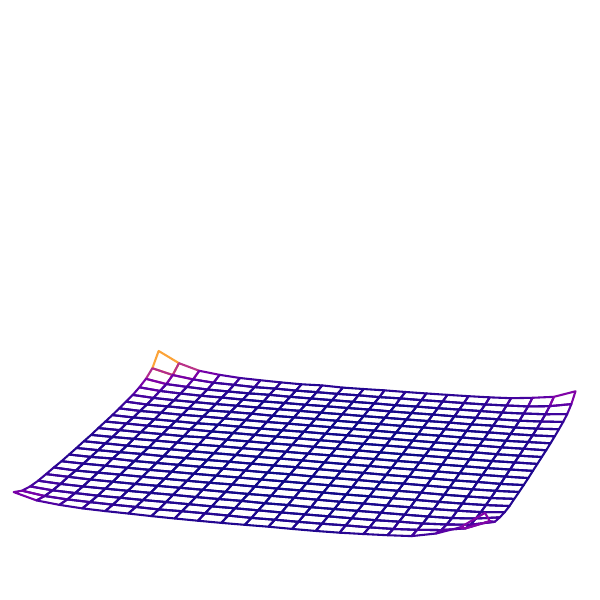}
            \caption*{TE2}
        \end{subfigure}
        \begin{subfigure}[t]{0.24\textwidth}
            \includegraphics[width=\textwidth]{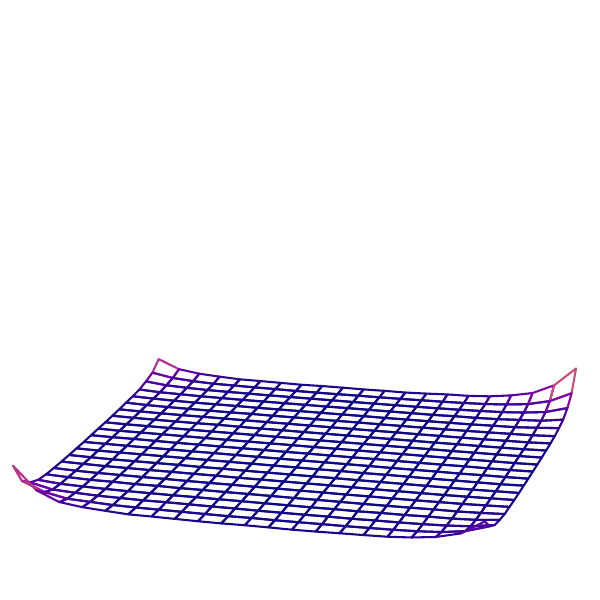}
            \caption*{TE3}
        \end{subfigure}
        \caption{KAdaptation~\cite{he2022parameter}}
    \end{subfigure} \\
    \begin{subfigure}[t]{1.0\textwidth}
        \begin{subfigure}[t]{0.24\textwidth}
            \includegraphics[width=\textwidth]{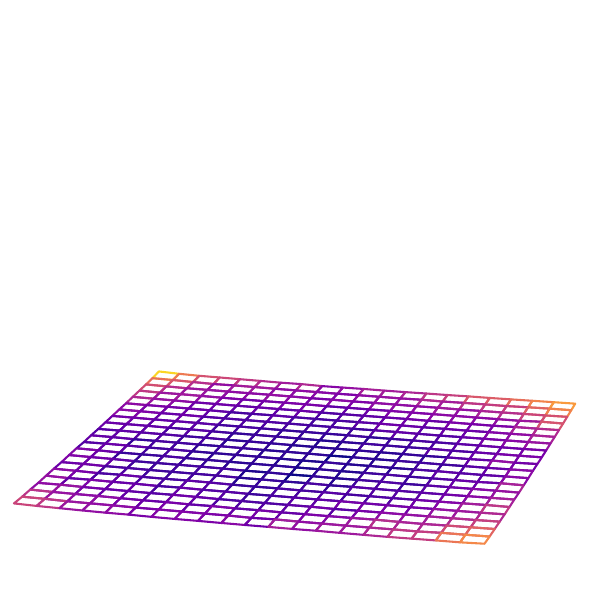}
            \caption*{TE0}
        \end{subfigure}
        \begin{subfigure}[t]{0.24\textwidth}
            \includegraphics[width=\textwidth]{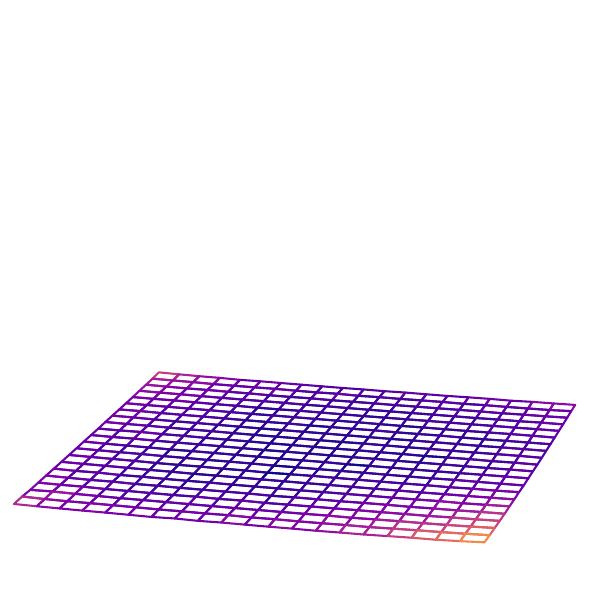}
            \caption*{TE1}
        \end{subfigure}
        \begin{subfigure}[t]{0.24\textwidth}
            \includegraphics[width=\textwidth]{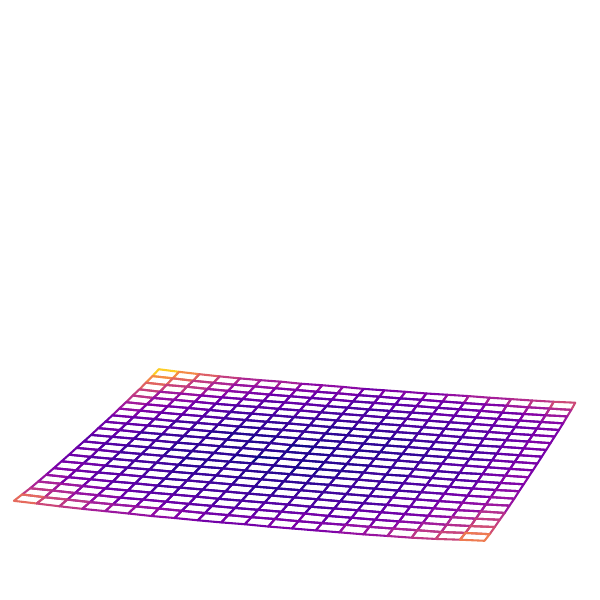}
            \caption*{TE2}
        \end{subfigure}
        \begin{subfigure}[t]{0.24\textwidth}
            \includegraphics[width=\textwidth]{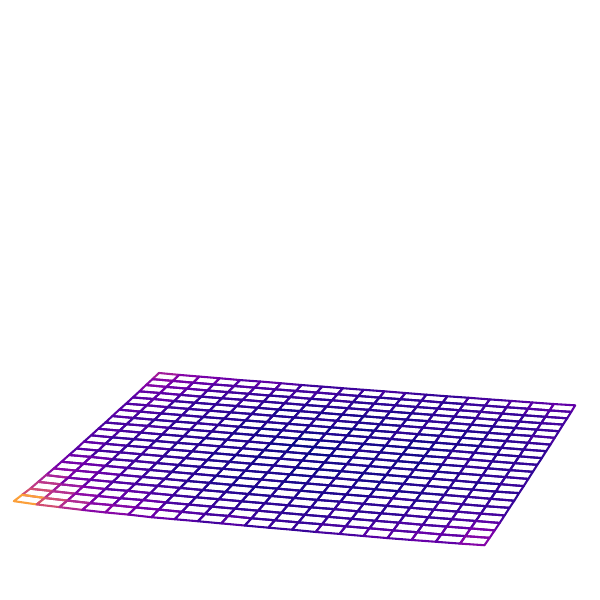}
            \caption*{TE3}
        \end{subfigure}
        \caption{KAdaptation with Mixture-of-Adapter (Ours)}
    \end{subfigure} \\
    \caption{Flatness comparison of loss surfaces trained with full fine-tuning, LoRA, KAdaptation, and KAdaptation with mixture-of-expert on the PACS dataset~\cite{li2017deeper}.}
    \label{fig:appendix-loss-landscape}
\end{figure*}

We show loss landscapes for all test environments in PACS dataset~\cite{li2017deeper} in Fig.~\ref{fig:appendix-loss-landscape}. Similar with the visualizations in main paper, the other test environments have a tendency that fully fine-tuned models show most sharp loss landscape. But trained models with LoRA and KAdaptation shows much more flatter loss landscapes, especially KAdaptation have most flat loss landscape.

\subsection{Analysis about the effectiveness of auxiliary loss}
\begin{figure*}[!]
    \centering
    \begin{subfigure}[t]{1.0\textwidth}
        \begin{subfigure}[t]{0.455\textwidth}
            \includegraphics[width=\textwidth]{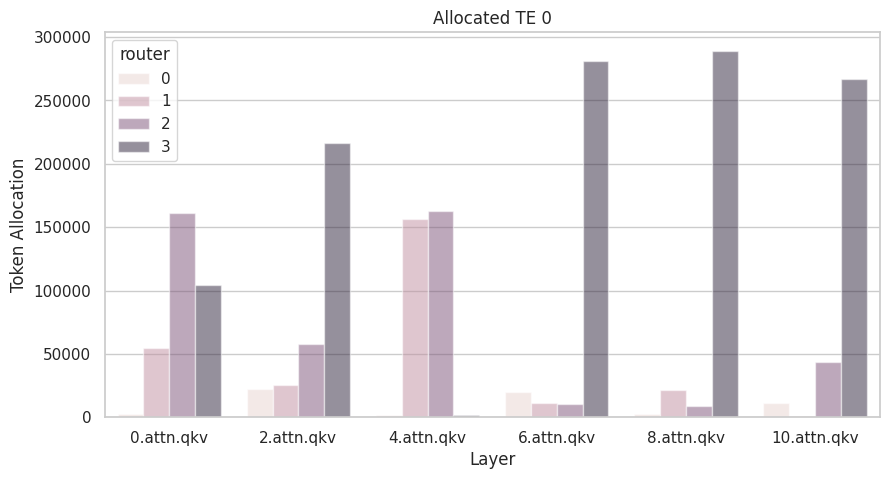}
        \end{subfigure}\hfill
        \begin{subfigure}[t]{0.49\textwidth}
            \includegraphics[width=\textwidth]{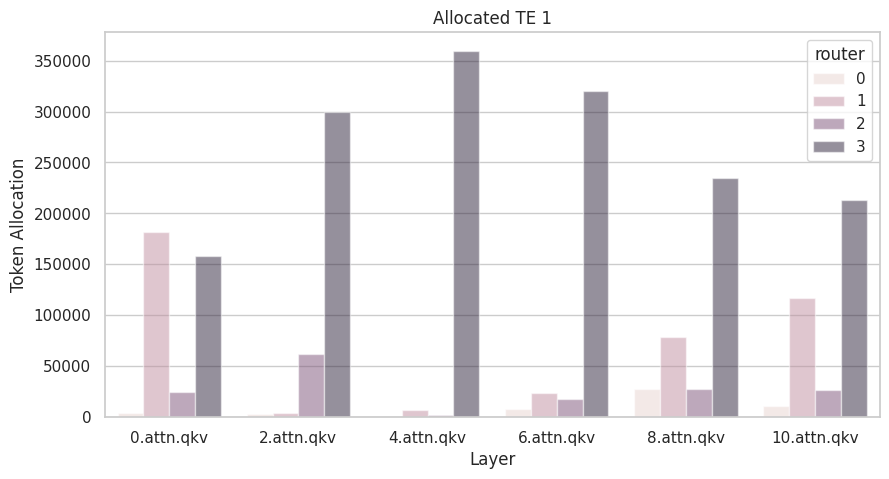}
        \end{subfigure} \\
        \begin{subfigure}[t]{0.455\textwidth}
            \includegraphics[width=\textwidth]{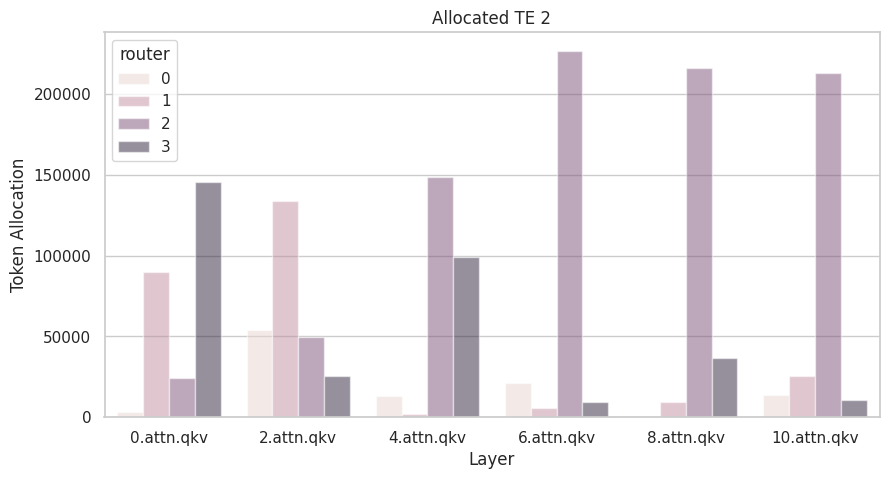}
        \end{subfigure}\hfill
        \begin{subfigure}[t]{0.49\textwidth}
            \includegraphics[width=\textwidth]{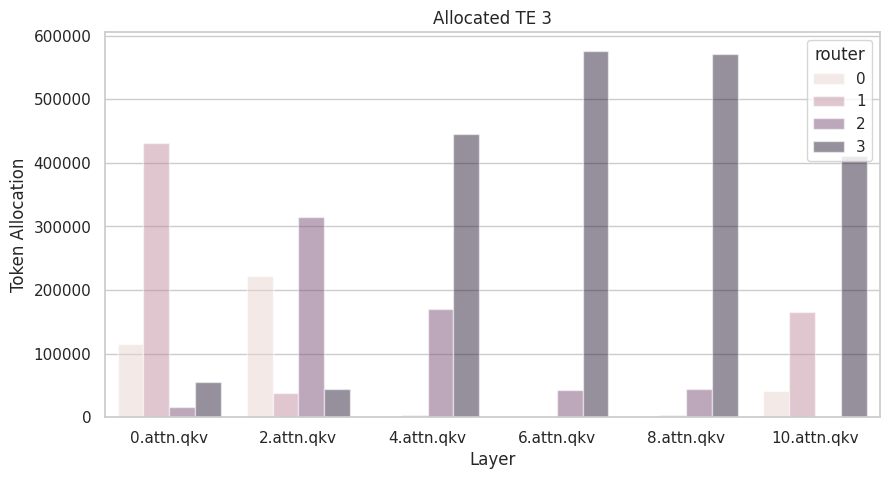}
        \end{subfigure} \\
    \vspace{5pt}
    \caption*{Without $\mathcal{L}_\mathrm{aux}$}
    \end{subfigure} \\
    \begin{subfigure}[t]{1.0\textwidth}
        \begin{subfigure}[t]{0.455\textwidth}
            \includegraphics[width=\textwidth]{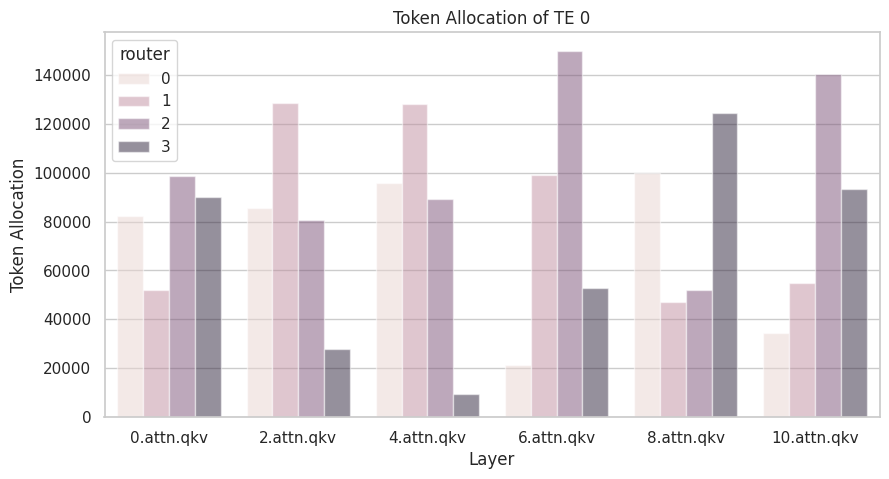}
        \end{subfigure}\hfill
        \begin{subfigure}[t]{0.49\textwidth}
            \includegraphics[width=\textwidth]{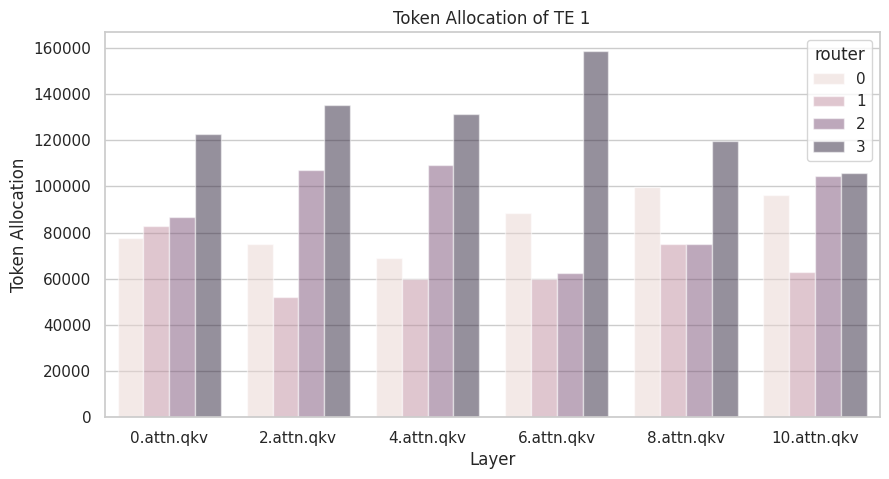}
        \end{subfigure} \\
        \begin{subfigure}[t]{0.455\textwidth}
            \includegraphics[width=\textwidth]{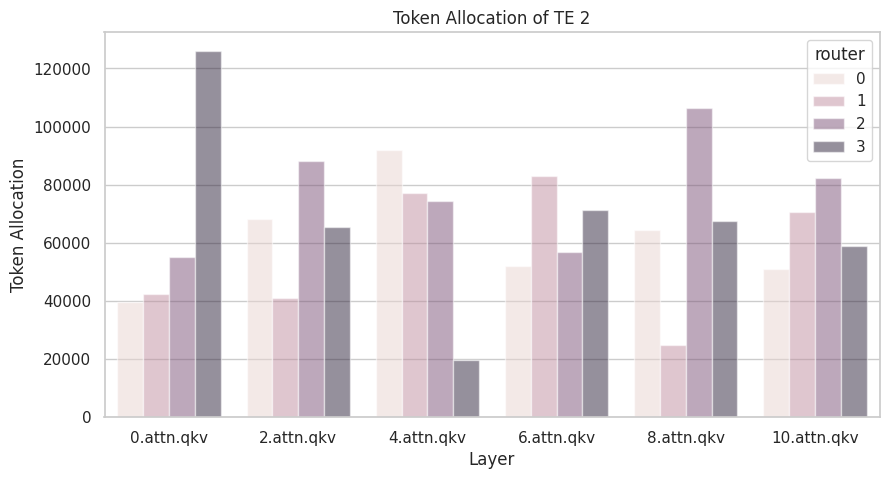}
        \end{subfigure}\hfill
        \begin{subfigure}[t]{0.49\textwidth}
            \includegraphics[width=\textwidth]{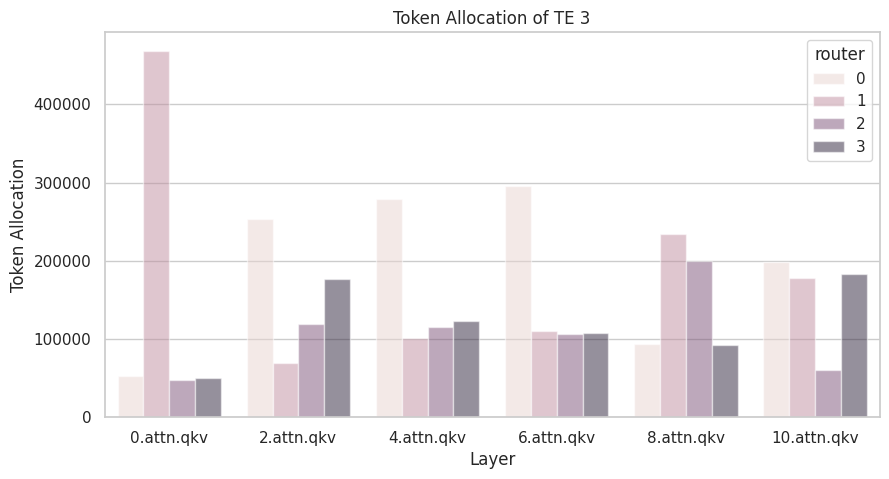}
        \end{subfigure} \\
    \vspace{5pt}
    \caption*{With $\mathcal{L}_\mathrm{aux}$}
    \end{subfigure} \\
    \caption{Visualizations of token routing tendencies with and without the auxiliary loss on PACS dataset. TE0 to TE3 correspond to the domains in the PACS dataset: Art\_painting, Cartoon, Photo, and Sketch. The x-axis represents the layer names containing the router and experts, while the y-axis shows the number of tokens allocated to each expert.}
    \label{fig:appendix-laux-routing}
\end{figure*}

In this section, we analyze how our model’s router allocates each token according to $\mathcal{L}_\mathrm{aux}$. As shown in Fig.~\ref{fig:appendix-laux-routing}, without the auxiliary loss, the router’s token allocation to the experts is highly imbalanced. However, when the auxiliary loss is applied, the allocation becomes significantly more balanced. We show the standard deviation of the tokens in Table~\ref{table:laux}. The results indicate that training with $\mathcal{L}_\mathrm{aux}$ leads to a more balanced distribution of tokens across the experts. This balance could play a crucial role when scaling up the model or applying it to downstream tasks.

\subsection{Visualizations of routed patches in PACS and TerraIncognita dataset.}
\label{appendix:router}
\begin{figure*}[t]
\centering
\begin{subfigure}{0.23\textwidth}
  \centering
  \includegraphics[width=1\linewidth]{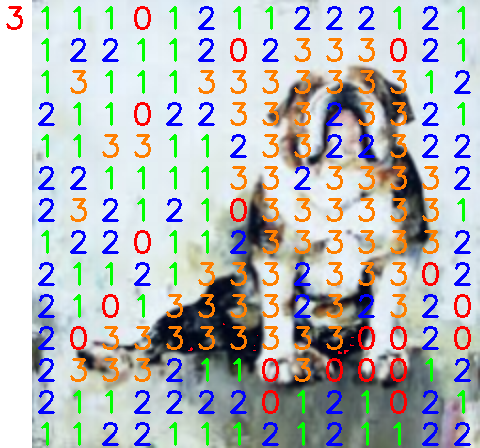}
\end{subfigure}
\begin{subfigure}{0.23\textwidth}
  \centering
  \includegraphics[width=1\linewidth]{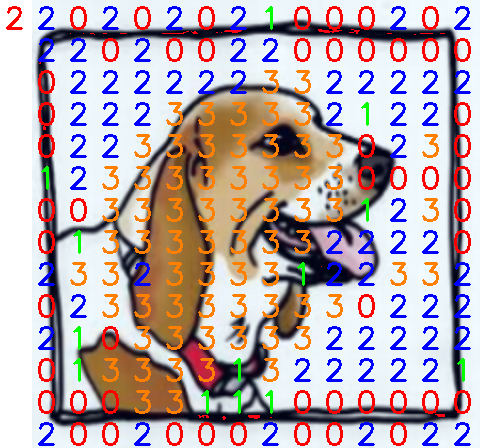}
\end{subfigure}
\begin{subfigure}{0.23\textwidth}
  \centering
  \includegraphics[width=1\linewidth]{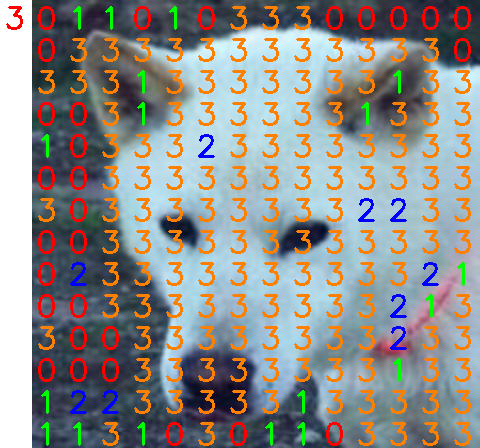}
\end{subfigure}
\begin{subfigure}{0.23\textwidth}
  \centering
  \includegraphics[width=1\linewidth]{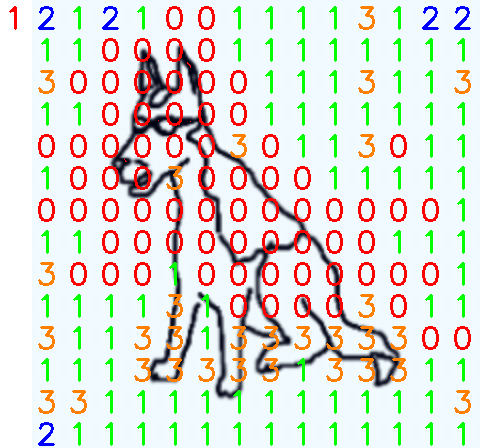}
\end{subfigure}\\
\begin{subfigure}{0.23\textwidth}
  \centering
  \includegraphics[width=1\linewidth]{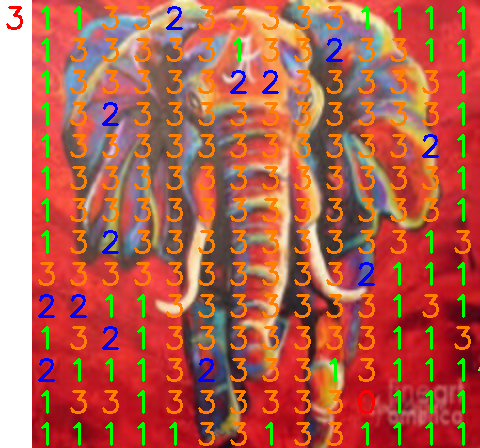}
\end{subfigure}
\begin{subfigure}{0.23\textwidth}
  \centering
  \includegraphics[width=1\linewidth]{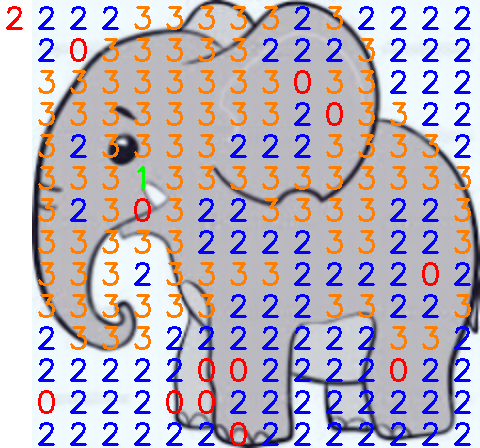}
\end{subfigure}
\begin{subfigure}{0.23\textwidth}
  \centering
  \includegraphics[width=1\linewidth]{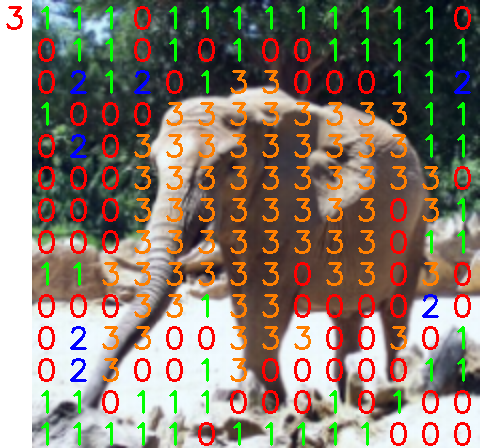}
\end{subfigure}
\begin{subfigure}{0.23\textwidth}
  \centering
  \includegraphics[width=1\linewidth]{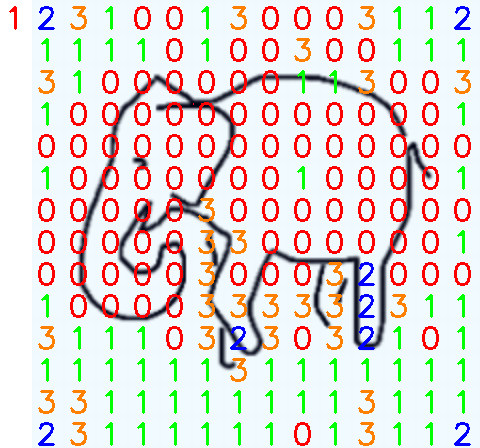}
\end{subfigure}\\
\begin{subfigure}{0.23\textwidth}
  \centering
  \includegraphics[width=1\linewidth]{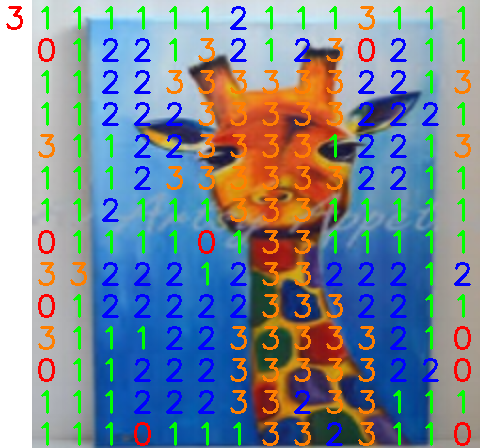}
\end{subfigure}
\begin{subfigure}{0.23\textwidth}
  \centering
  \includegraphics[width=1\linewidth]{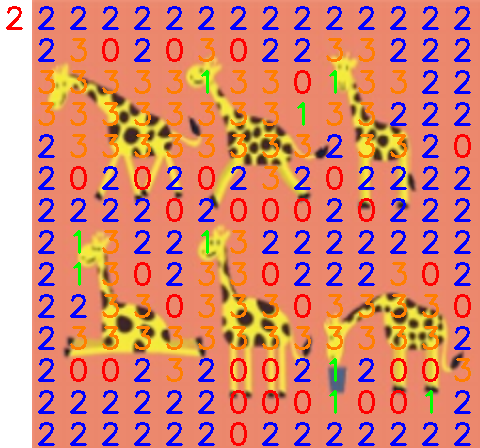}
\end{subfigure}
\begin{subfigure}{0.23\textwidth}
  \centering
  \includegraphics[width=1\linewidth]{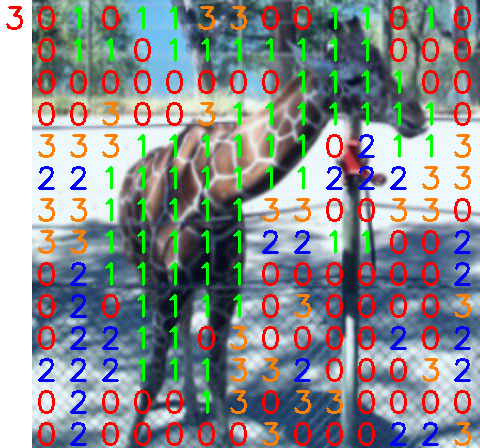}
\end{subfigure}
\begin{subfigure}{0.23\textwidth}
  \centering
  \includegraphics[width=1\linewidth]{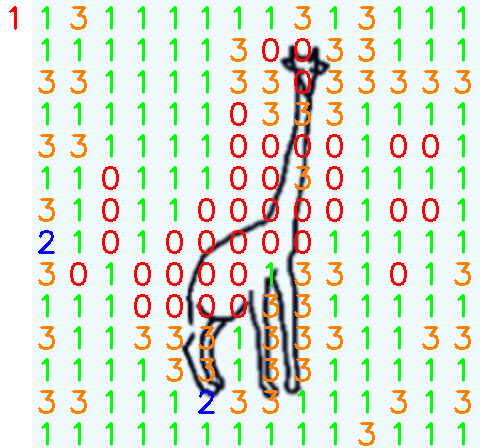}
\end{subfigure}\\
\vspace{-5pt}
\caption{Visualizations of routed indices of each patch. We show a total of seven classes in PACS dataset~\cite{li2017deeper}, with one class per row in the order of `Dog', `Elephant', `Giraffe'. Also, in each column, the same domains are located in the order of `Art Painting', `Cartoon', `Photo', and `Sketch'. }
\label{fig:routed-all1}
\end{figure*}

\begin{figure*}[t]
\centering
\begin{subfigure}{0.23\textwidth}
  \centering
  \includegraphics[width=1\linewidth]{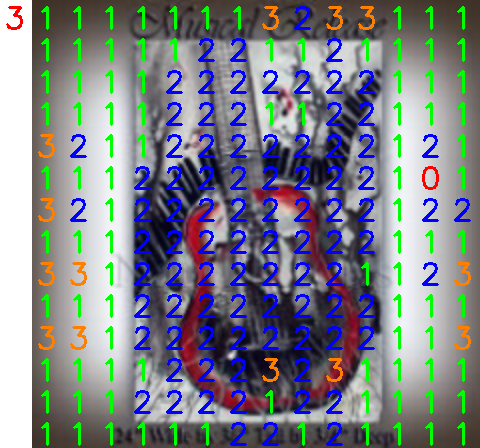}
\end{subfigure}
\begin{subfigure}{0.23\textwidth}
  \centering
  \includegraphics[width=1\linewidth]{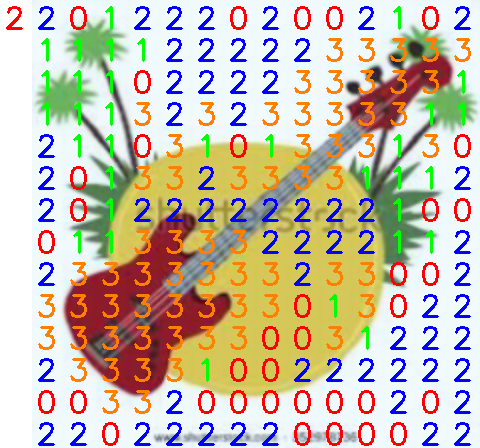}
\end{subfigure}
\begin{subfigure}{0.23\textwidth}
  \centering
  \includegraphics[width=1\linewidth]{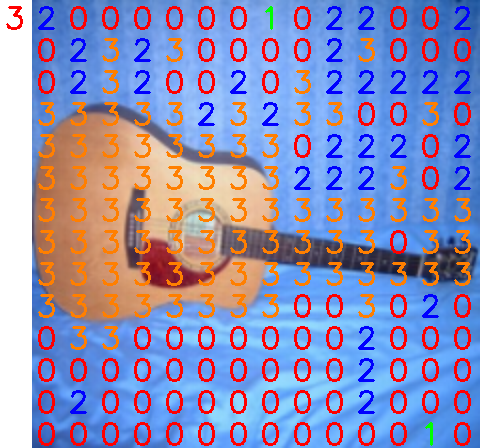}
\end{subfigure}
\begin{subfigure}{0.23\textwidth}
  \centering
  \includegraphics[width=1\linewidth]{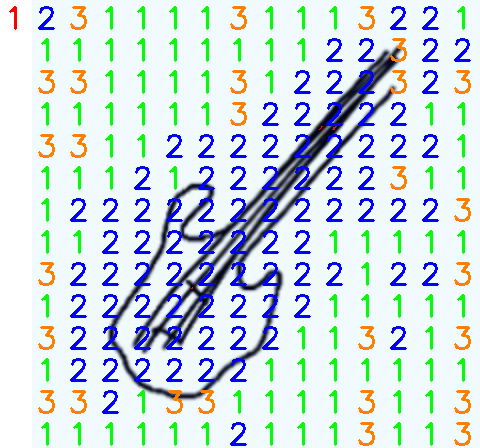}
\end{subfigure}\\
\begin{subfigure}{0.23\textwidth}
  \centering
  \includegraphics[width=1\linewidth]{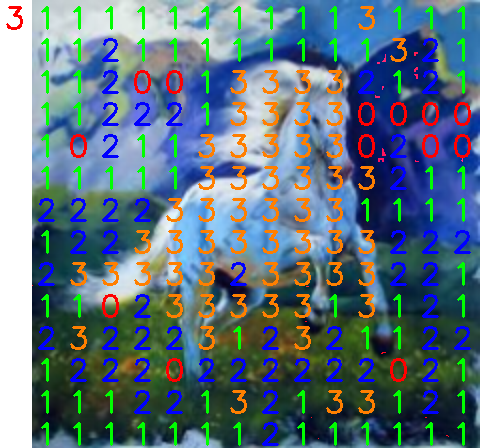}
\end{subfigure}
\begin{subfigure}{0.23\textwidth}
  \centering
  \includegraphics[width=1\linewidth]{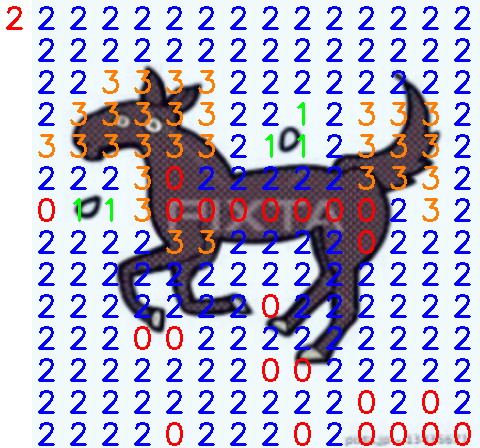}
\end{subfigure}
\begin{subfigure}{0.23\textwidth}
  \centering
  \includegraphics[width=1\linewidth]{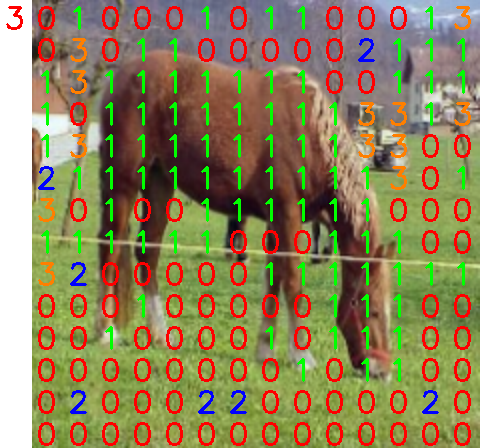}
\end{subfigure}
\begin{subfigure}{0.23\textwidth}
  \centering
  \includegraphics[width=1\linewidth]{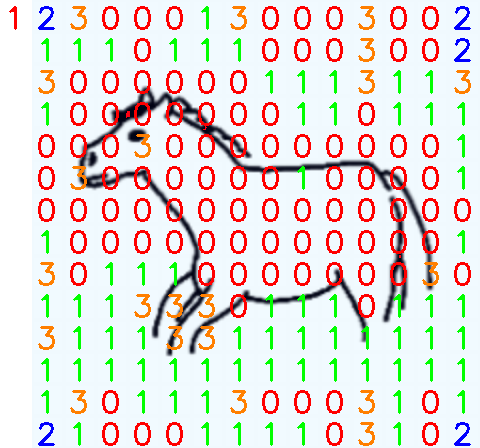}
\end{subfigure}\\
\begin{subfigure}{0.23\textwidth}
  \centering
  \includegraphics[width=1\linewidth]{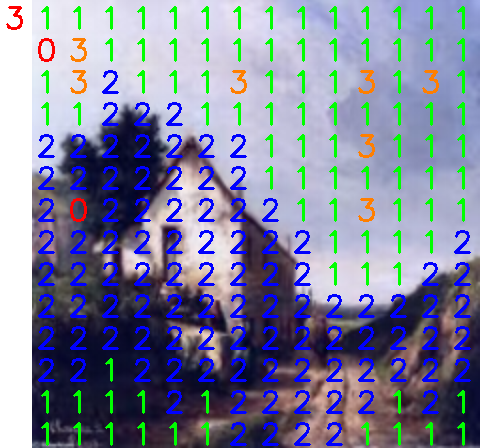}
\end{subfigure}
\begin{subfigure}{0.23\textwidth}
  \centering
  \includegraphics[width=1\linewidth]{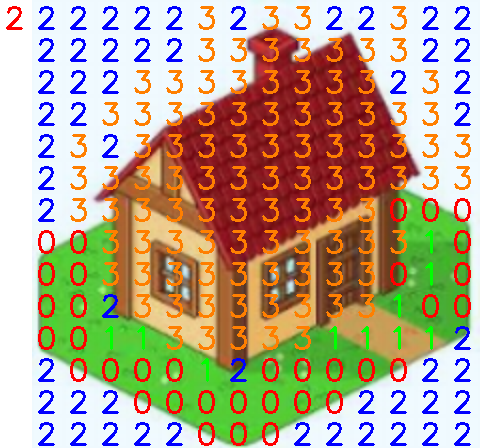}
\end{subfigure}
\begin{subfigure}{0.23\textwidth}
  \centering
  \includegraphics[width=1\linewidth]{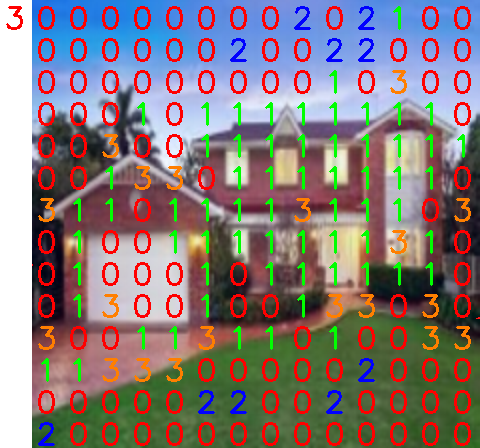}
\end{subfigure}
\begin{subfigure}{0.23\textwidth}
  \centering
  \includegraphics[width=1\linewidth]{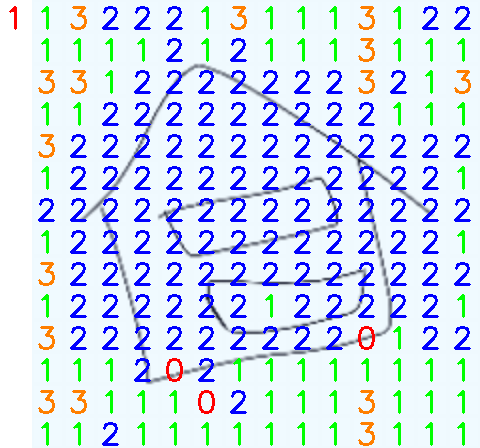}
\end{subfigure}\\
\begin{subfigure}{0.23\textwidth}
  \centering
  \includegraphics[width=1\linewidth]{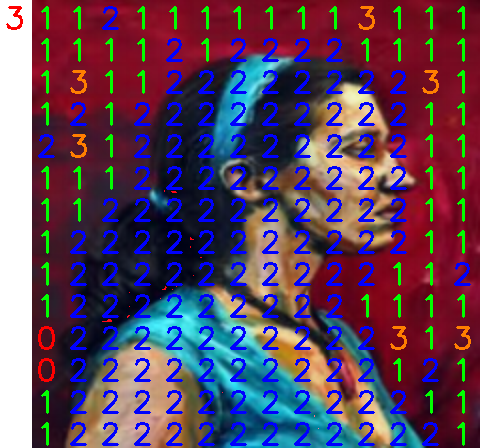}
\end{subfigure}
\begin{subfigure}{0.23\textwidth}
  \centering
  \includegraphics[width=1\linewidth]{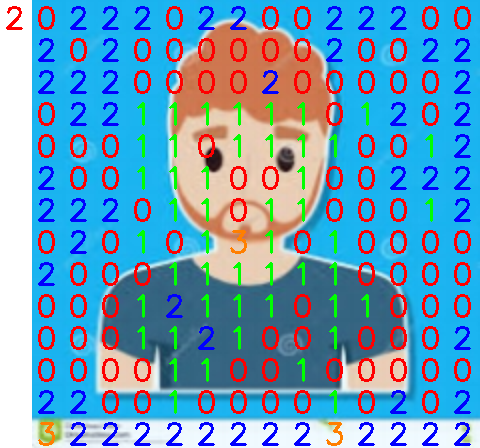}
\end{subfigure}
\begin{subfigure}{0.23\textwidth}
  \centering
  \includegraphics[width=1\linewidth]{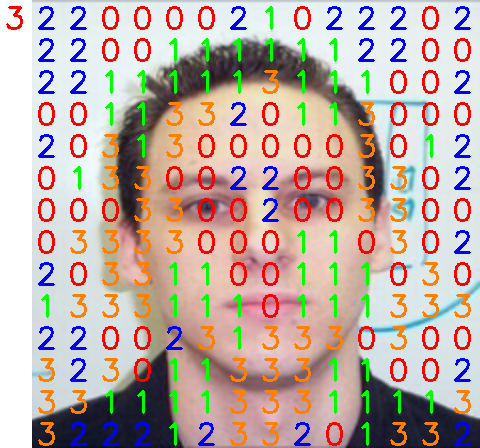}
\end{subfigure}
\begin{subfigure}{0.23\textwidth}
  \centering
  \includegraphics[width=1\linewidth]{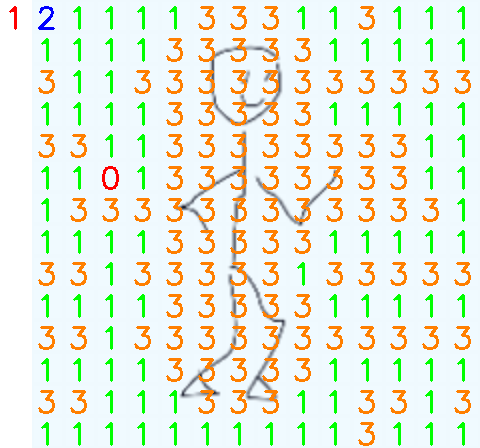}
\end{subfigure}
\vspace{-5pt}
\caption{Visualizations of routed indices of each patch. We show a total of seven classes in PACS dataset~\cite{li2017deeper}, with one class per row in the order of `Guitar', `Horse', `House', `Person'. Also, in each column, the same domains are located in the order of `Art Painting', `Cartoon', `Photo', and `Sketch'. }
\label{fig:routed-all2}
\end{figure*}

\begin{figure*}[!]
    \centering
    \begin{subfigure}[t]{1.0\textwidth}
        \begin{subfigure}[t]{0.455\textwidth}
            \includegraphics[width=\textwidth]{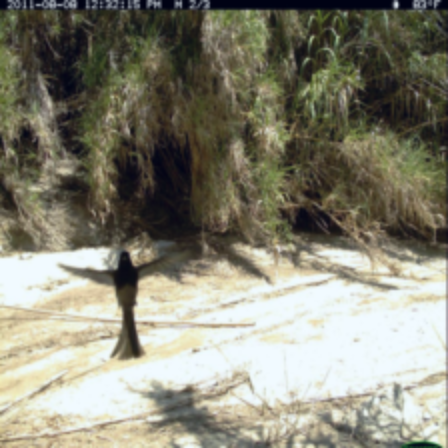}
        \end{subfigure} \hfill
        \begin{subfigure}[t]{0.49\textwidth}
            \includegraphics[width=\textwidth]{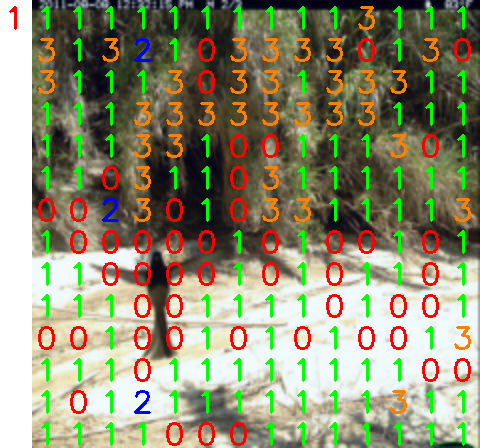}
        \end{subfigure} \\
        \begin{subfigure}[t]{0.455\textwidth}
            \includegraphics[width=\textwidth]{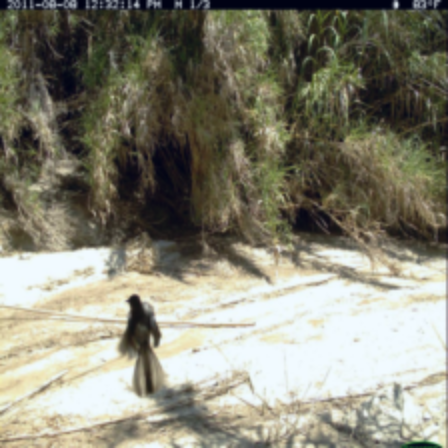}
        \end{subfigure}\hfill
        \begin{subfigure}[t]{0.49\textwidth}
            \includegraphics[width=\textwidth]{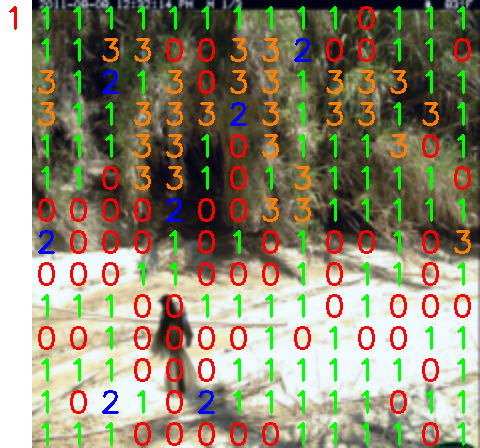}
        \end{subfigure}\\
    \vspace{5pt}
    \caption*{Location 43}
    \end{subfigure}
    \caption{Visualizations of routed indices for each patch in the TerraIncognita~\cite{beery2018recognition} dataset. The left column displays the original image, while in the right column, we indicate where each patch is routed. The upper and lower images were taken at the same location but different times, therefore they share the same background but feature different object (bird) in terms of shape and location.}
    \label{fig:appendix-routing-terra1}
\end{figure*}

\begin{figure*}[!]
    \centering
    \begin{subfigure}[t]{1.0\textwidth}
        \begin{subfigure}[t]{0.455\textwidth}
            \includegraphics[width=\textwidth]{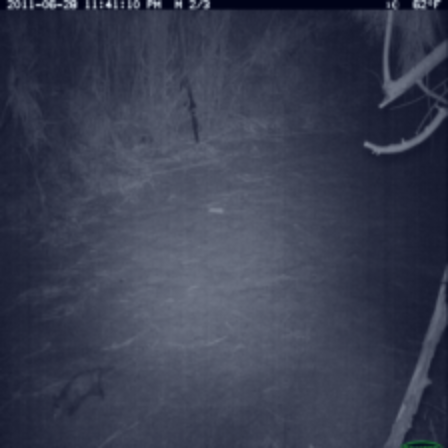}
        \end{subfigure}\hfill
        \begin{subfigure}[t]{0.49\textwidth}
            \includegraphics[width=\textwidth]{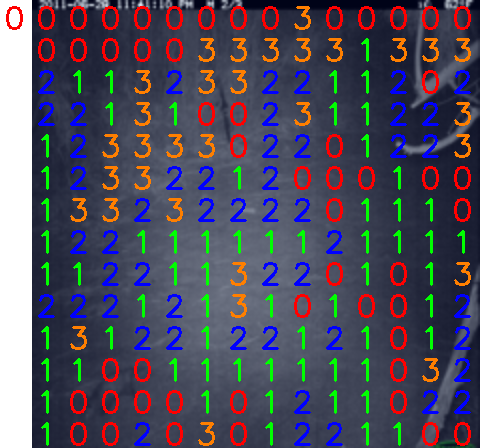}
        \end{subfigure} \\
        \begin{subfigure}[t]{0.455\textwidth}
            \includegraphics[width=\textwidth]{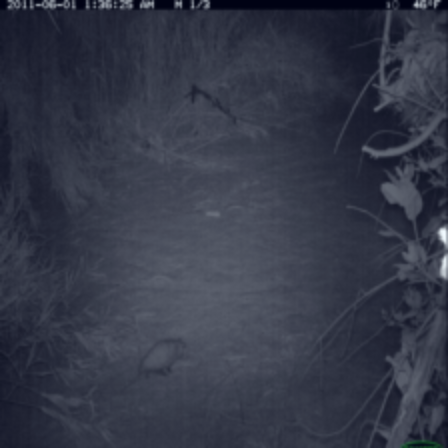}
        \end{subfigure}\hfill
        \begin{subfigure}[t]{0.49\textwidth}
            \includegraphics[width=\textwidth]{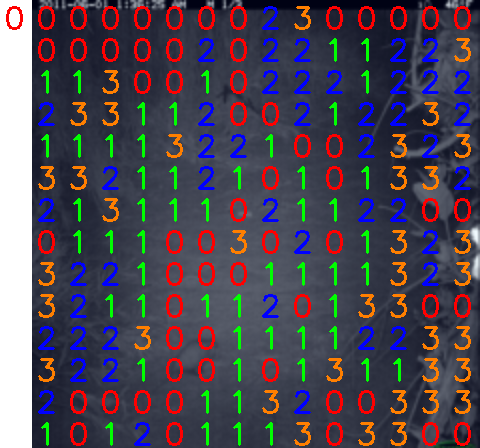}
        \end{subfigure} \\
    \vspace{5pt}
    \caption*{Location 46}
    \end{subfigure} \\
    \caption{Visualizations of routed indices for each patch in the TerraIncognita~\cite{beery2018recognition} dataset. The left column displays the original image, while in the right column, we indicate where each patch is routed. The upper and lower images were taken at the same location but different times, therefore they share the same background but feature different object (opossum) in terms of shape and location.}
    \label{fig:appendix-routing-terra2}
\end{figure*}

We additionally show the visualizations of routed patch indices in Fig.~\ref{fig:routed-all1},~\ref{fig:routed-all2} on PACS dataset~\cite{li2017deeper}, and Fig.~\ref{fig:appendix-routing-terra1},~\ref{fig:appendix-routing-terra2} on TerraIncognita dataset~\cite{beery2018recognition}. All images are visualizations from the last adapter-attached transformer layer, layer 10. Similar with the findings from main paper, we can observe that same indices are clustered at the regions where having semantic meanings.


\subsection{Limitations}
Our method heavily relies on the performance of large pretrained models, hence using a better pretrained model can lead to improved performance. But, such models are limited and require a substantial amount of time and cost for training. These weakness also exist in methods like MIRO~\cite{cha2022domain} or SIMPLE~\cite{li2023simple}, and the availability of high-performance open-source models like OpenCLIP~\cite{ilharco_gabriel_2021_5143773} can alleviate these drawbacks. Our approach may not significantly outperform on datasets more challenging than TerraIncognita due to fewer trainable parameters compared to fully fine-tuned DG algorithms. However, it offers flexibility by adjusting trainable parameters via the inner rank $r_i$, and optimal rank can be obtained through hyperparameter search, effectively addressing this limitation.
\end{multicols}

\end{document}